\documentclass{article} 
\usepackage{iclr2025_conference,times}

\usepackage{amsmath,amsfonts,bm}

\def\eqref#1{equation~(\ref{#1})}
\def\Eqref#1{Equation~(\ref{#1})}

\def\1{\bm{1}}

\DeclareMathAlphabet{\mathsfit}{\encodingdefault}{\sfdefault}{m}{sl}
\SetMathAlphabet{\mathsfit}{bold}{\encodingdefault}{\sfdefault}{bx}{n}

\usepackage{amsmath}
\usepackage{amssymb}
\usepackage{mathtools}
\usepackage{amsthm}
\usepackage{algorithm}
\usepackage{algorithmic,wrapfig}
\newtheorem{proposition}{Proposition}
\newtheorem{definition}{Definition}
\usepackage[hidelinks,pagebackref=true]{hyperref}
\usepackage{url}
\usepackage{todonotes}
\newcommand{\prob}{\ensuremath{\mathbb{P}}}

\title{How does Your RL Agent Explore?\\ An Optimal Transport Analysis of Occupancy Measure Trajectories}

\author{Reabetswe M.~Nkhumise\textsuperscript{1}, Debabrota Basu\textsuperscript{2}, Tony J.~Prescott\textsuperscript{1}, Aditya Gilra\textsuperscript{3,1} \\
\textsuperscript{1} Department of Computer Science,
The University of Sheffield, UK.\\
~~~\texttt{\{rmnkhumise1,t.j.prescott\}@sheffield.ac.uk} \\
\textsuperscript{2}
\'Equipe Scool, Univ. Lille, Inria, CNRS, Centrale Lille, UMR 9189 CRIStal, France.\\
~~~\texttt{debabrota.basu@inria.fr} \\
\textsuperscript{3}
Centrum Wiskunde \& Informatica,
Amsterdam, Netherlands.\\
~~~\texttt{aditya.gilra@cwi.nl} \\
}

\iclrfinalcopy 
\begin{document}

\maketitle

\begin{abstract}
The rising successes of RL are propelled by combining smart algorithmic strategies and deep architectures to optimize the distribution of returns and visitations over the state-action space. 
A quantitative framework to compare the learning processes of these eclectic RL algorithms is currently absent but desired in practice.  
We address this gap by representing the learning process of an RL algorithm as a sequence of policies generated during training, and then studying the policy trajectory induced in the manifold of state-action occupancy measures. 
Using an optimal transport-based metric, we measure the length of the paths induced by the policy sequence yielded by an RL algorithm between an initial policy and a final optimal policy. 
Hence, we first define the \textit{Effort of Sequential Learning (ESL)}. ESL quantifies the relative distance that an RL algorithm travels compared to the shortest path from the initial to the optimal policy. 
Further, we connect the dynamics of policies in the occupancy measure space and regret (another metric to understand the suboptimality of an RL algorithm), by defining the \textit{Optimal Movement Ratio} (OMR). OMR assesses the fraction of movements in the occupancy measure space that effectively reduce an analogue of regret. 
Finally, we derive approximation guarantees to estimate ESL and OMR with finite number of samples and without access to an optimal policy.
Through empirical analyses across various environments and algorithms, we demonstrate that ESL and OMR provide insights into the exploration processes of RL algorithms and hardness of different tasks in discrete and continuous MDPs.
\end{abstract}

\section{Introduction}\label{sec:intro}

In recent years, significant advancements in Reinforcement Learning (RL) have been achieved in developing exploration techniques that improve learning~\citep{Bellemare16,Burda18,Eysenbach19} along with new learning methods~\citep{Lazaridis20,Muller21,Li23}. 
With growing computational resources, these techniques have led to various successful applications of RL, such as playing games up to human proficiency~\citep{Silver17,Jaderberg19}, controlling robots~\citep{Ibarz21,Kaufmann23}, tuning databases and computer systems~\citep{Wang21,Basu19},  etc.
However, there remains a lack of consensus over approaches that can quantitatively compare these exploratory processes across RL algorithms and tasks~\citep{VanSeijen20,Amin21,Ladosz22}. This is attributed to some methods being algorithm-specific~\citep{Tang17}, while others provide theoretical guarantees for very specific settings~\citep{Lattimore20,Agarwal22}. 
Thus, comparing the exploratory processes of these eclectic algorithms across the multi-directional space of RL algorithm design, emerges as a natural question. However, the present literature lacks a metric to compare them except regret, which is often hard to estimate~\citep{Ramos17,Ramos18}.



This paper aims to address this gap based on two key observations.
\textit{First}, we observe from the linear programming formulation of RL that solving the value maximization problem is equivalent to finding an optimal occupancy measure~\citep{Syed08,Neu20,Kalagarla21}.
Occupancy measure is the distribution of state-action pair visits induced by a policy~\citep{Altman99,Laroche23}. Under mild assumptions, a policy maps uniquely to an occupancy measure.
\textit{Second}, we observe that any RL algorithm learns by sequentially updating policies starting from an initial policy to reach an optimal policy. 
The search for an optimal policy is influenced by the exploration-exploitation strategy and functional approximators, both of which impact the overall performance of the agent by determining the quality of experiences from which it learns~\citep{Zhang19,Ladosz22}. 
Hereby, we term collectively the learning strategy and the exploration-exploitation interplay as the \textit{exploratory process}.

\textbf{Contributions.} \textit{1. A Framework.} Motivated by our observations, we abstract any RL algorithm as a trajectory of occupancy measures induced by a sequence of policies between an initial and a final (optimal) policy.
The occupancy measure of a policy given an environment corresponds to the data-generating distribution of state-actions. Thus, we can quantify the effort of each policy update, i.e. the effort to shift the state-action data distributions, as the transportation distance between their occupancy measures. The total effort of learning by the algorithm can be measured as the total distance covered by its occupancy measure trajectory.
We provide a mathematical basis for this quantification by proving that the space of occupancy measures is a differentiable manifold for smoothly parameterized policies (Section~\ref{sec:framework}). 
Hence, we can compute the length of the occupancy measure trajectory on this manifold using Wasserestein distance as the metric~\citep{Villani09}.



\textit{2. Effort of Sequential Learning.} In contrast to RL, if we knew the optimal policy we could update our initial policy directly via supervised or imitation learning. Effort of this learning is represented by a direct, shortest (geodesic) path from initial to optimal policy on the occupancy measure manifold. 
To quantify the cost of the exploratory process to learn the environment, we define the \textit{Effort of Sequential Learning} (ESL) as the ratio of the (indirect) path traversed by an RL algorithm in the occupancy measure space to the direct distance between the initial and optimal policy (Section~\ref{sec:esl}). Lower ESL implies more efficient exploratory process.

\textit{3. Efforts to learn that lead to Regret-analogue minimization.} Regret is a widely used optimality measures for reward-maximizing RL algorithms~\citep{Sutton18}. It measures the total deviation in the value functions achieved by a sequence of policies learned by an RL algorithm with
respect to the optimal algorithm that always uses the optimal policy~\citep{Sinclair23}. We show that regret is related to the sum of distances between the optimal policy and each policy in the sequence learned by the RL algorithm, in the occupancy measure space. We can define an analogue of instantaneous regret (at any one step during learning rather than cumulative), in the occupancy measure space, as the geodesic distance between the occupancy measure of the policy at this step in the learning sequence, and the optimal one. We find that not all policy updates lead to a reduction in this analogue of immediate regret, and thus define another index \textit{Optimal Movement Ratio} that measures the fraction that do (Section~\ref{sec:omr}).

\textit{4. Computational and Numerical Insights.} We prove sample complexity guarantees to approximate ESL and OMR in practice as we do not have access to the occupancy measures but collection of rollouts from the corresponding policies (Section~\ref{sec:computational}). We show the relation of empirical OMR and ESL to the true ones if the optimal policy is never reached by an algorithm. We conduct experiments on multiple environments, both discrete and continuous, with sparse and dense rewards, comparing state-of-the-art algorithms. We observe that by visualizing aspects of the path traversed (and by comparing ESL and OMR), we are able to compare and provide insights into their exploratory processes and the impact of task hardness on them (Section~\ref{sec:experiments}). The results confirm the ubiquity and effectiveness of our approach to study the exploratory processes of RL algorithms.


\section{Preliminaries}

\textbf{Markov Decision Processes.} Consider an agent interacting with an environment in discrete timesteps. At each timestep ${t} \in \mathbb{N}$, the agent observes a state ${s_t}$, executes an action ${a_t}$, and receives a scalar reward ${\mathcal{R}(s_t,a_t)}$. The behaviour of the agent is defined by a policy ${\pi}(a_t|s_t)$, which maps the observed states to actions. The environment is modelled as a Markov Decision Process (MDP) $\mathbb{M}$ with a state space $\mathcal{S}$, action space $\mathcal{A}$, transition dynamics $\mathcal{T}: \mathcal{S}\times \mathcal{A} \to \mathcal{S}$, and reward function ${\mathcal{R}: \mathcal{S} \times \mathcal{A} \to \mathbb{R}}$. During task execution, the agent issues actions in response to states visited, and hence a sequence of states and actions ${h_{t} = (s_0,a_0,s_1,a_1,...,s_{t-1},a_{-1},s_t)}$, here called a rollout, is observed.

In infinite-horizon settings, the state value function for a given policy $\pi$ is the expected discounted cumulative reward over time ${V_{\pi}(s) \triangleq \mathbb{E}_{\pi} \left[ \sum_{t=0}^{\infty} \gamma^{t} \mathcal{R}(s_t,a_t) \mid s_0 = s\right]}$, where ${\gamma \in [0,1)}$ is the discount rate. The goal is to learn a policy that maximises the objective ${J^{\pi}_{\mu} \triangleq \mathbb{E}_{s \sim \mu} [V_{\pi}(s)]}$, where $\mu(s)$ is the initial state distribution.

\textbf{Occupancy Measure.} 
The state-action occupancy measure is a distribution over the ${\mathcal{S} \times \mathcal{A}}$ space that represents the discounted frequency of visits to each state-action pair when executing a policy ${\pi}$ in the environment \citep{Syed08}. Formally, the occupancy measure of $\pi$ is $v_{\pi}(s,a) \triangleq \rho \sum_{t=0}^{\infty} \gamma^{t} \prob(s_{t} = s, a_{t} = a \mid \pi, \mu)$, where $\rho = 1- \gamma$ is the normalizing factor.

Stationary Markovian policies allow a bijective correspondence with their state-action occupancy measures \citep{Givchi21}. We express the objective $J^{\pi}_{\mu}$ in terms of the occupancy measure as 
\begin{equation} \label{objective_occupancy}
J^{\pi}_{\mu} = \frac{1}{\rho} \mathbb{E}_{(s,a) \sim v_{\pi}} \left[ \mathcal{\bar{R}}(s,a) \right], 
\end{equation}
where $\mathcal{\bar{R}}(s,a)$ is the expected immediate reward for the state-action pair $(s,a)$. 

\textbf{Wasserstein Distance.} Let $\mu,\nu \in \mathcal{P}(\mathcal{X})$ be probability measures on a complete and separable metric (Polish) space $(\mathcal{X},d_{\mathcal{X}})$. The p-Wasserstein distance between $\mu$ and $\nu$ is~\citep{Villani09}
\begin{equation}\label{wasserstein_distance}
\begin{aligned}
    \mathcal{W}_{p}(\mu,\nu) \triangleq \left( \min_{\pi \in \Pi(\mu,\nu)} \int_{\mathcal{X} \times \mathcal{X}} c(x,x') \,d\pi(x,x') \right)^{1/p}\,,
\end{aligned}
\end{equation}
where the cost function is given by the metric as $c(x,x') = (d_{\mathcal{X}}(x,x'))^{p}$ for some $p \geq 1$. ${\Pi(\mu,\nu)}$ is a set of all admissible transport plans between $\mu$ and $\nu$, i.e. probability measures on $\mathcal{X} \times \mathcal{X}$ space with marginals $\mu$ and $\nu$. We consider the 1-Wasserstein distance where $p = 1$. 

\textbf{MDPs with Lipschitz Rewards.} 
Following \cite{Pirotta15} and \cite{Kallel24}, we assume an MDP with $L_{\mathcal{R}}$-Lipschitz rewards (ref. Appendix \ref{Appendix Lipschitz MDP} for elaboration) that satisfies 
    $|\mathcal{\bar{R}}(s,a) - \mathcal{\bar{R}}(s',a')| \leq L_{\mathcal{R}}d_{\mathcal{S}\mathcal{A}}((s,a),(s',a'))$ 
    for all $s, s' \in \mathcal{S}$ and $a, a' \in \mathcal{A}$.
Here, $d_{\mathcal{S}\mathcal{A}}((s,a),(s',a')) = d_{\mathcal{S}}((s,s')) + d_{\mathcal{A}}((a,a'))$ is the metric defined on the joint state-action space $\mathcal{S} \times \mathcal{A}$. 
This is a weaker condition than assuming a completely Lipschitz MDP. \citet{Pirotta15} showed that for any pair of stationary policies $\pi$ and $\pi'$, the absolute difference between their corresponding objectives is 
\begin{equation}\label{performance_distance_bound}
    \left| J^{\pi}_{\mu} - J^{\pi'}_{\mu} \right| \leq \frac{L_{\mathcal{R}}}{\rho}\mathcal{W}_{1}(v_{\pi},v_{\pi'})\,,
\end{equation}
where $\mathcal{W}_{1}(v_{\pi},v_{\pi'})$ is the 1-Wasserstein distance between the occupancy measures of the policies (ref. Appendix~\ref{Appendix  Performance Difference} for details). 


\section{RL Algorithms as Trajectories of Occupancy Measures}\label{sec:framework}
The exploration process (i.e. the exploration-exploitation interplay and learning strategy) of an RL algorithm, influence how the policy model updates its policies~\citep{Kaelbling96,Sutton18}. During training, a \textit{policy trajectory}, i.e. sequence of policies $(\pi_{0},\pi_{1}, \dots, \pi_{N})$, is generated during policy updates due to the exploratory process. We assume these policies belong to a set of stationary Markov policies parameterised by $\theta$. For policies in this set $\pi_{\theta} \in \mathbf{\Gamma}_{\theta}$, we define the space of occupancy measures corresponding to $\mathbf{\Gamma}_{\theta}$ as $\mathcal{M} = \{ v_{\pi_{\theta}}(s,a) \mid \pi_{\theta} \in \mathbf{\Gamma}_{\theta}, \theta \in \mathbb{R}^{N_\theta} \}$. 

\begin{proposition}[Properties of $\mathcal{M}$]\label{prop:M_manifold}
If the policy $\pi$ has a smooth parameterization $\theta$ and the inverse of $P^{\pi}(s,s') \triangleq \sum_{a}T(s \mid s',a)\pi(a \mid s')$ exists, then the space of occupancy measures $\mathcal{M}$ is a differentiable manifold. 
(Proof in Appendix~\ref{Appendix prop 1}) 
\end{proposition}

Thus, we can compute the length of any path on this differentiable manifold $\mathcal{M}$ using an appropriate metric. The manifold $\mathcal{M}$ of occupancy measures is endowed with a 1-Wasserstein metric to compute the distance between occupancy measures corresponding to policies parameterized by $\theta, \theta+d\theta \in \mathcal{M}$ as $ds = \mathcal{W}_{1}(v_{\pi_{\theta}},v_{\pi_{\theta+d\theta}})$. In imitation learning, the 1-Wasserstein distance between the occupancy measures of the learner and expert can be used as a minimizable loss function to learn the expert's policy~\citep{Zhang20}. Hence, the 1-Wasserstein distance reflects the effort required to achieve this imitation learning. Similarly, we propose the following quantification for the effort to update from one policy to another.

\begin{definition}[Effort of Learning]\label{eol}
    We define the 1-Wasserstein metric between occupancy measures of two policies $\pi$ and $\pi'$, i.e. $\mathcal{W}_{1}(v_{\pi},v_{\pi'})$, as the effort required to learn or update from one policy to the other.
\end{definition}

Therefore, when a learning process causes an update between occupancy measures in $\mathcal{M}$, we attribute the resulting update effort to the learning process and refer to it as the effort of learning. In a learning process, first the initial policy $\pi_0$ is obtained typically by randomly sampling the model parameters, then these parameters $\theta$ undergo updates until a predefined convergence criterion is satisfied, yielding the final optimal policy $\pi_N=\pi^*$. Since each policy has a corresponding occupancy measure, this process yields a sequence of points on $\mathcal{M}$, which can be connected by geodesics between successive points, producing a curve. The length of the curve is computed by the summation of the finite geodesic distances between consecutive policies along it~\citep{lott08} 
\begin{equation} \label{curve_length}
    C \triangleq \sum_{k=0}^{N-1}  \mathcal{W}_{1}(v_{\pi_{\theta_{k}}},v_{\pi_{\theta_{k+1}}})\,,
\end{equation}
where $\theta_{0}$ and $\theta_{N}$ are respectively the initial and final parameter values before and after learning.

\subsection{Effort of Sequential Learning (ESL)}\label{sec:esl}
\begin{figure}[t!]
\begin{minipage}{0.48\textwidth}
\centering\includegraphics[width=0.75\columnwidth]{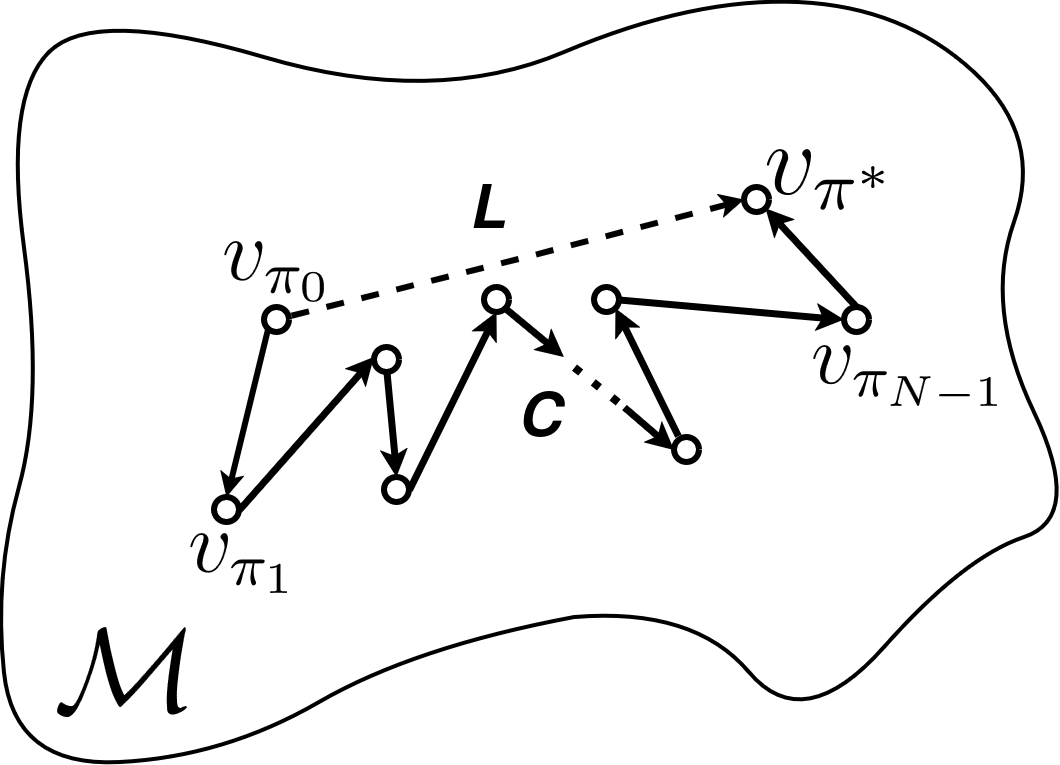}
\caption{ Schematic of the policy trajectory $C$ in the space of occupancy measures $\mathcal{M}$ during RL training (solid line) vs. the geodesic $L$ (shortest path, dashed line) between the initial and final points (i.e. $\pi_{0}$ and $\pi_{N} = \pi^{*} $).}\label{trajectory_depiction}
\end{minipage}\hfill
\begin{minipage}{0.5\textwidth}\vspace*{-1.5em}
\centering\includegraphics[width=0.9\columnwidth]{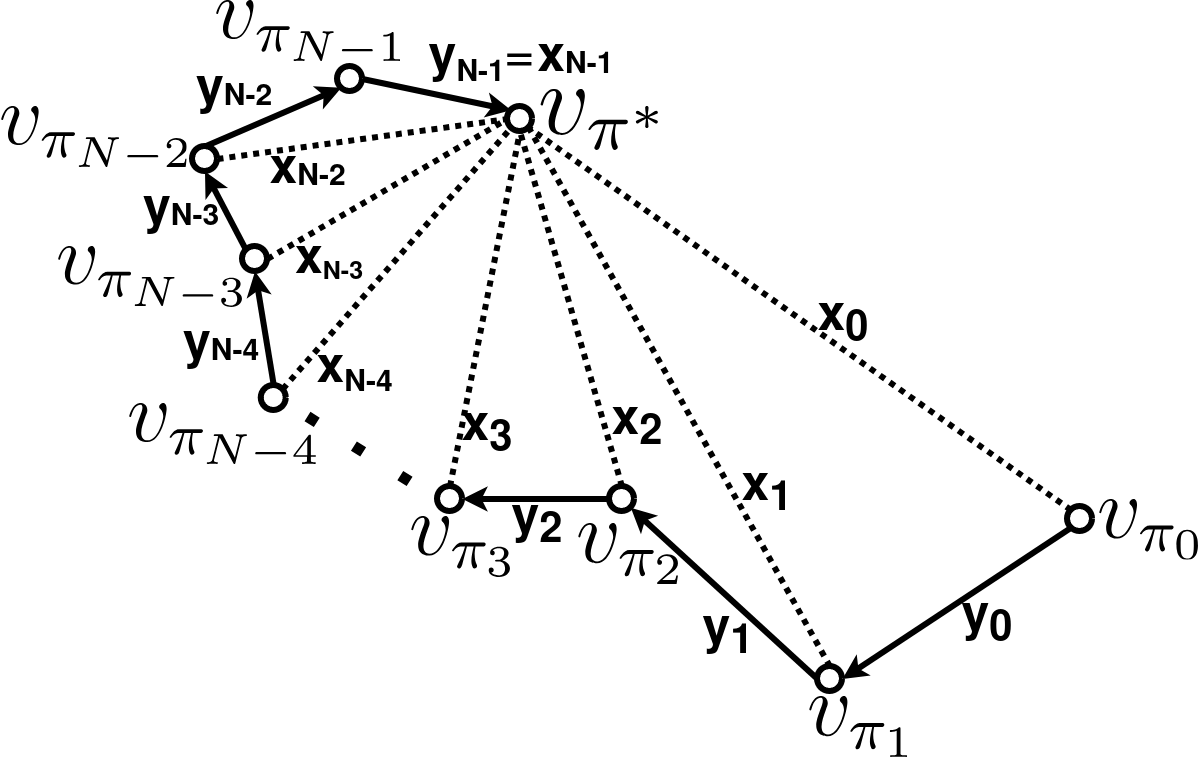}\vspace*{-.5em}
\caption{Schematic of how \textit{distance-to-optimal} (denoted by $x_{k}$) and \textit{stepwise-distance} (denoted by $y_{k}$) on the occupancy measure space describe exploratory process of an RL algorithm during training.}\label{distance_depiction}\vspace*{-1.5em}
\end{minipage}
\end{figure}

As we saw above, RL generates a trajectory in the occupancy measure manifold $\mathcal{M}$, whose length is given by Equation~(\ref{curve_length}). Compared to the long trajectory of sequential policies generated by the exploratory process, the geodesic $L$ is the ideal, shortest path to the optimal policy $\pi_N = \pi^*$ from $\pi_{0}$, whose length is $L = \mathcal{W}_{1}(v_{\pi_{0}},v_{\pi_{N}})$. This path would be taken by an imitation-learning oracle algorithm that knows $\pi^*$. Both these paths are schematically depicted in Figure \ref{trajectory_depiction}.

\begin{definition}[Effort of Sequential Learning (ESL)]\label{defn:esl}
    We define the effort of sequential learning incurred by a trajectory of the exploratory process of an RL algorithm, relative to the oracle that knows $\pi^* (=\pi_N)$, as 
\begin{equation}\label{sequential_effort}
    \eta \triangleq \frac{\sum_{k=0}^{N-1}  \mathcal{W}_{1}(v_{\pi_{k}},v_{\pi_{k+1}})}{\mathcal{W}_{1}(v_{\pi_{0}},v_{\pi_{N}})} 
\end{equation}
Due to the stochasticity of the exploratory process, we introduce an expectation to obtain $\bar{\eta} = \mathbb{E}_{\pi_{0}, \mu} \left[ \mathrm{\eta} \right]$. We refer to $\bar{\eta}$ as the \textit{effort of sequential learning} (ESL). 
\end{definition}

Note that $\bar{\eta} \geq 1$, and a larger $\bar{\eta}$ corresponds to a less efficient exploratory process of the RL algorithm. Hence, an RL algorithm with $\bar{\eta} \approx 1$ closely mimics the oracle and has an efficient exploratory process. 

\subsection{Optimal Movement Ratio (OMR)}\label{sec:omr}
Regret measures the total deviation in value functions incurred by a sequence of policies learned by an RL algorithm with respect to the optimal algorithm that always uses the optimal policy \citep{Sinclair23}. We show that regret is connected to the sum of distances from each policy in the sequence learned by an RL algorithm to the optimal policy in the occupancy measure space.
\begin{proposition}[Regret and Occupancy Measures]\label{prop:regret}
Given an MDP with $L_{\mathcal{R}}$-Lipschitz rewards, we obtain
    $\text{Regret} \triangleq \sum_{k=1}^N \left(   J^{\pi^*}_{\mu} - J^{\pi_k}_{\mu} \right) \leq \frac{L_{\mathcal{R}}}{\rho}\sum_{k=1}^N \mathcal{W}_{1}(v_{\pi_{k}},v_{\pi^*})$. (Proof in Appendix \ref{Appendix prop 2})
\end{proposition}


We refer to $\mathcal{W}_{1}(v_{\pi_{k}},v_{\pi^*})$ as the \textit{distance-to-optimal}, and analogously use it as the expected immediate regret in the occupancy measure space. Furthermore, we refer to $\mathcal{W}_{1}(v_{\pi_{k}},v_{\pi_{k+1}})$ as \textit{stepwise-distance}. Interestingly, during training, the \textit{distance-to-optimal} and \textit{stepwise-distance} share a relationship illustrated in Figure~\ref{distance_depiction}. From Figure~\ref{distance_depiction}, we observe that if the change in \textit{distance-to-optimal}, $\delta_{k} \triangleq \mathcal{W}_{1}(v_{\pi_{k}},v_{\pi^*}) - \mathcal{W}_{1}(v_{\pi_{k+1}},v_{\pi^*}) > 0$, it indicates that the agent got closer to the optimal. We define the set $K^+$ as containing indices $k$ for which $\delta_k > 0$, while $K^-$ contains the rest. 

\begin{definition}[Optimal Movement Ratio (OMR)]\label{defn:omr}
We define the proportion of policy transitions that effectively reduce the \textit{distance-to-optimal}, in a learning trajectory, as
\begin{equation} \label{optimal_movement_ratio}
\begin{aligned}
   \kappa &\triangleq \frac{ \sum_{k \in K^+}\mathcal{W}_{1}(v_{\pi_{k}},v_{\pi_{k+1}}) }{ \sum_{k=0}^{N-1} \mathcal{W}_{1}(v_{\pi_{k}},v_{\pi_{k+1}})}\,.
\end{aligned}
\end{equation}
Due to the stochasticity of the exploratory process, we introduce an expectation to obtain $\bar{\kappa} = \mathbb{E}_{\pi_{0}, \mu} \left[ \kappa \right]$. We refer to $\bar{\kappa}$ as the \textit{optimal movement ratio} (OMR). 
\end{definition}
Note that $\bar{\kappa} \in [0,1]$, and $\bar{\kappa} \to 1$ indicates that nearly all the policy updates reduce the \textit{distance-to-optimal}, thus showing high efficiency. $\bar{\kappa} \to 0$ implies low efficiency, since only a small fraction of the policy updates contribute towards the reduction of the \textit{distance-to-optimal}.

\subsection{Extension to Finite-Horizon Episodic Setting}\label{sec:finite_horizon}
In the episodic finite-horizon MDP formulation of RL, in short \textit{Episodic RL}~\citep{Osband13, Azar17, Ouhamma23}, the agent interacts with the environment in multiple episodes of $H$ steps. An episode starts by observing state $s_1$. Then, for $t=1,\ldots H$, the agent draws action $a_t$ from a (possibly time-dependent) policy $\pi_t(\cdot\mid s_t)$, observes the reward $r(s_t,a_t)$, and transits to a state $s_{t+1} \sim T(\cdot\mid s_t,a_t)$. Here, the value function and the state-action value functions at step $h\in [H]$ are defined as $V_h^\pi (s) \triangleq \mathbb{E}_{\mathbb{M},\pi}\left[\sum_{t=h}^H r(s_t,a_t) \mid s_h =s \right]$, and $Q_h^\pi (s,a) \triangleq \mathbb{E}_{\mathbb{M},\pi}\left[\sum_{t=h}^H r(s_t,a_t) \mid s_h =s, a_h=a \right]$.
Following~\citep{Altman99}, we can define a finite-horizon version of occupancy measures as 
\begin{equation} \label{episodic_occupancy}
\begin{aligned}
    v^H_{\pi}(s,a) \triangleq \frac{1}{H}\sum_{t=1}^{H} \prob(s_{t} = s, a_{t} = a \mid \pi, \mu).
\end{aligned}
\end{equation}

Following~\citep{Syed08}, we can show that $v^H_{\pi}$ satisfies the linear programming description of value function maximization along with the Bellman flow constraints (ref. Sec II.C. in~\cite{Kalagarla21}). Additionally, we prove that under some assumptions, the finite-horizon occupancy measures also construct a manifold, referred as $\mathcal{M}^H$.
\begin{proposition}[Properties of $\mathcal{M}^H$]\label{prop:M_H_manifold}
If the policy $\pi$ has a smooth parametrization $\theta$ and the inverses of both $(\mathbb{I}- P^{\pi})$ and $ P^{\pi}$ exist, then the space of finite-horizon occupancy measures $\mathcal{M}^H$ is a differentiable manifold. (Proof in Appendix~\ref{Appendix prop 3}) 
\end{proposition}
This allows us to similarly define a Wasserstein metric on this manifold, which in turn, allows us to compute ESL and OMR to evaluate different RL algorithms.

\section{Computational Challenges and Solutions}\label{sec:computational}
Similar to regret, our method requires knowing the optimal policy. This is because the efficiency and effectiveness of exploratory processes of RL algorithms are highly coupled with their ability to reach optimality. ESL and OMR depend on the policies being stationary and Markovian. 

\subsection{Policy datasets for computing occupancy measures}
We consider approximations of occupancy measures using datasets assumed to be drawn from these measures. We estimate the Wasserstein distance between the occupancy measures using a method introduced by \cite{Alvarez20} known as the \textit{optimal transport dataset distance} (OTDD). OTDD uses datasets to estimate the Wasserstein distance between the underlying distributions. See Appendix~\ref{Appendix OTDD} for a detailed account on OTDD. 

\textbf{Definition 4 }(Policy dataset). \textit{A dataset of a policy $\mathcal{D}_{\pi}$ is a set of state-action pairs drawn from the policy's occupancy measure, i.e. $\mathcal{D}_{\pi} = \{ (s_{(i)},a_{(i)}) \}_{i=1}^{m} \sim v_{\pi}$. These can be constituted from the rollouts generated by the policy during task execution.}

We know from imitation learning literature that if we are given $\mathcal{D}_{\pi}$, generated by an expert policy, we can train a policy model on it in a supervised manner via behaviour cloning~\citep{Hussein17}. Thus, knowing $\mathcal{D}_{\pi}$ can allow converting an RL task into a Supervised Learning (SL) task. Consider a scenario when we have access to a sequence of datasets $(\mathcal{D}_{\pi_0}, \dots ,\mathcal{D}_{\pi_N})$, each corresponding to policy $\pi_t$ for ${t \geq 0}$. If we train (in a supervised manner) a policy model sequentially on these datasets, the model will undergo a similar policy evolution as via the RL algorithm that generated the policy trajectory $(\pi_t)_{t \geq 0}$. This allows us to conceptualise learning in RL as a sequence of SL tasks with  sequential transfer learning across the datasets $(\mathcal{D}_{\pi_0}, \dots ,\mathcal{D}_{\pi_N})$. We employ OTDD to estimate $W_{1}( v_{\pi_{k}},v_{\pi_{k+1}})$ using these datasets, i.e. $d_{OT}(\mathcal{D}_{\pi_{k}},\mathcal{D}_{\pi_{k+1}}) \approx W_{1}( v_{\pi_{k}},v_{\pi_{k+1}})$, based on Proposition~\ref{prop:pac}.


\begin{proposition}[Upper Bound on Estimation Error]\label{prop:pac}
Let an RL algorithm yield a sequence of policies $\pi_0, \ldots ,\pi_N$ while training. Now, we construct $N$ datasets $\mathcal{D}_{\pi_0}, \dots ,\mathcal{D}_{\pi_N}$, each consisting of $M$ rollouts of the corresponding policies. Then, we can use these datasets to approximate $\sum_{k=0}^{N-1}  \mathcal{W}_{1}(v_{\pi_{\pi_{k}}},v_{\pi_{\pi_{k+1}}})$ by $\sum_{k=0}^{N-1}  d_{OT}(\mathcal{D}_{\pi_{k}},\mathcal{D}_{\pi_{k+1}})$ with an expected error upper bound $\frac{2 N\mathcal{E}_{2}} {\sqrt{M}}+ N\gamma^{T+1}diam(\mathcal{SA})$.
Here, $T$ is the total number of steps per episode, $diam(\mathcal{SA})$ is the diameter of the state-action space, and $\mathcal{E}_{2}$ is a positive-valued and polylogarithmic function of $S$ and $A$. For finite horizon case, we can further reduce the error bound to $\frac{2 N\mathcal{E}_{2}} {\sqrt{M}}$.
\end{proposition}
Proof of Proposition~\ref{prop:pac} is in Appendix~\ref{Appendix prop 4}. The results support that ESL and OMR can be estimated as 
\begin{align}\label{sequential_effort_approx}
     \bar{\eta} &= \mathbb{E}_{\pi_{0}, \mu} \left[ \frac{\sum_{k=0}^{N-1}  d_{OT}(\mathcal{D}_{\pi_{k}},\mathcal{D}_{\pi_{k+1}})}{d_{OT}(\mathcal{D}_{\pi_{0}},\mathcal{D}_{\pi_{N}})} \right]\,, 
     ~\text{ and }~
     \bar{\kappa} = \mathbb{E}_{\pi_{0}, \mu} \left[ \frac{ \sum_{k \in K^+} d_{OT}(\mathcal{D}_{\pi_{k}},\mathcal{D}_{\pi_{k+1}}) }{ \sum_{k=0}^{N-1} d_{OT}(\mathcal{D}_{\pi_{k}},\mathcal{D}_{\pi_{k+1}})} \right]  \,.
\end{align} 

\subsection{When an optimal policy is not reached}
So far we have assumed that the algorithms converge at the optimal policy, i.e. $\pi_{N} = \pi^{*}$. However, this is not always true. We consider a scenario when $\pi_{N} \neq \pi^{*}$, and define  
\begin{equation} \label{non_optimal_sequential_effort}
\begin{aligned}
    \eta_{sub} = \frac{\sum_{k=0}^{N-1}  \mathcal{W}_{1}(v_{\pi_{\pi_{k}}},v_{\pi_{\pi_{k+1}}})}{\mathcal{W}_{1}(v_{\pi_{0}},v_{\pi_{N}})} , \pi_{N} \neq \pi^{*}\,.
\end{aligned}
\end{equation}
\begin{proposition}\label{prop:eta_sub}
Given $N\geq 2$ and $\pi_0 \neq \pi_N \neq \pi^*$, we obtain
\begin{equation} \label{sequential_effort_vs_non_optimal}
    \frac{\eta -\eta_{sub}}{\eta} \leq \frac{2\mathcal{W}_{1}(v_{\pi_{N}},v_{\pi^{*}})}{ \mathcal{W}_{1}(v_{\pi_{0}},v_{\pi_N})} \,.
\end{equation}
\end{proposition}
This is true due to the triangle inequalities: $\mathcal{W}_{1}(v_{\pi_{0}},v_{\pi^{*}}) + \mathcal{W}_{1}(v_{\pi_{N}},v_{\pi^{*}}) \geq \mathcal{W}_{1}(v_{\pi_{0}},v_{\pi_{N}})$ and $\mathcal{W}_{1}(v_{\pi_{N-1}},v_{\pi_{N}}) + \mathcal{W}_{1}(v_{\pi_{N}},v_{\pi^{*}}) \geq \mathcal{W}_{1}(v_{\pi_{N-1}},v_{\pi^{*}})$. 
Equation~(\ref{sequential_effort_vs_non_optimal}) shows that in the case where $\pi_{N}$ is close to $\pi^{*}$, then $\eta_{sub}$ is a good approximation of $\eta$, and thus, a good quantifier to determine the efficiency of the algorithm's exploratory process. The proof is in Appendix~\ref{Appendix Inequalities} and corresponding experimental results are in Appendix~\ref{Appendix B.2}.




\section{Experimental  Evaluation}\label{sec:experiments}
In this section, we evaluate the proposed methods in the \textit{2D-Gridworld} and \textit{Mountain Car} \citep{Moore90,Brockman16} environments, to analyse our methods in discrete and continuous state-action spaces respectively. The 2D-Gridworld environment is of size 5x5 with actions: \{up, right, down, left\}. In the gridworld, we perform experiments on 3 settings namely:- A) deterministic with dense rewards, B) deterministic with sparse rewards, and C) stochastic with dense rewards. Further details about these settings are provided in Appendix~\ref{Environment Description}. The Mountain Car environment, in our experimentation, is a deterministic MDP with dense rewards that consists of both continuous states and actions (described in detail in \citep{Brockman16}). The final experiment studies how ESL scales with task hardness in several gridworld environments of varying difficulty.


Our experiments aim to address the following questions:\newline
1. \textit{What information can the visualization of the policy evolution during RL training provide about the exploratory process of the algorithm?} \newline
2. \textit{How do ESL and OMR allow us to analyse the exploratory processes of RL algorithms?}\newline  
3. \textit{Does ESL scale proportionally with task difficulty?}

\textbf{Summary of Results.} In Section~\ref{trajectories_expo}, we demonstrate that visualizing evolution of \textit{distance-to-optimal} and \textit{stepwise-distance} of different RL algorithms during training reveals: 1) whether the agent is stuck in suboptimal policies, 2) the aggressiveness of the exploration processes, and 3) their tendencies. We further compare ESL and OMR of different algorithms on a few environments in Section~\ref{comparison_algos}. Finally, we show in Section~\ref{tasks_section} that ESL scales proportionally with task difficulty, and thus, reflects the effects of task difficulty on exploration and learning.

\vspace*{-1 em}\subsection{Exploration Trajectories of RL Algorithms}\label{trajectories_expo}
\begin{figure}[t!]
\centering\vspace*{-2em}
\setlength{\arrayrulewidth}{0.1mm}
\setlength{\tabcolsep}{0.0mm}

\begin{tabular}{llllll}
     \includegraphics[width=0.16\linewidth]{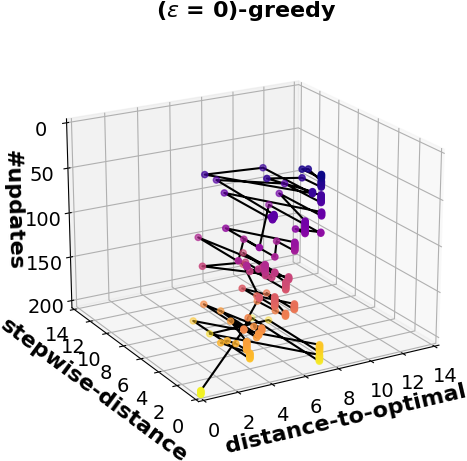} &  \includegraphics[width=0.16\linewidth]{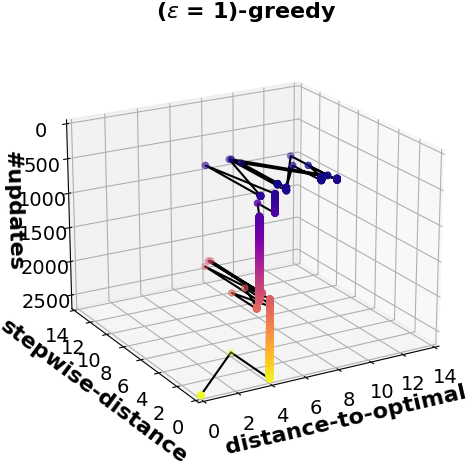} &  \includegraphics[width=0.16\linewidth]{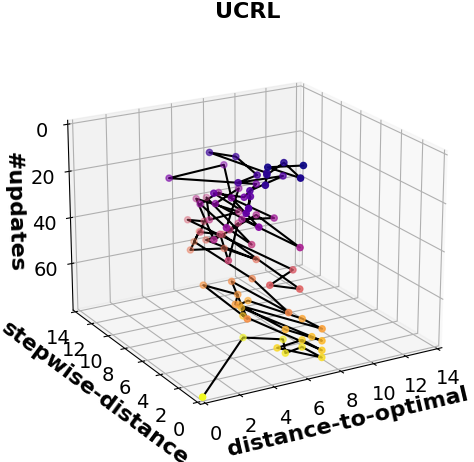} &  \includegraphics[width=0.16\linewidth]{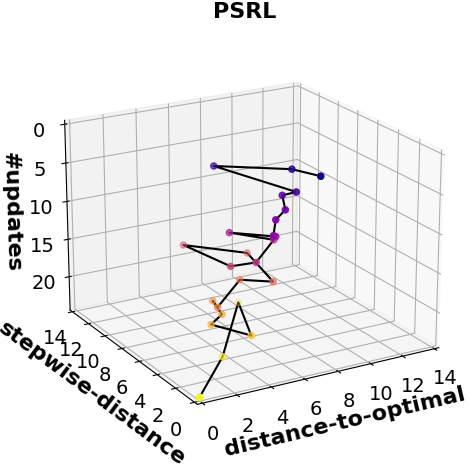} &  \includegraphics[width=0.16\linewidth]{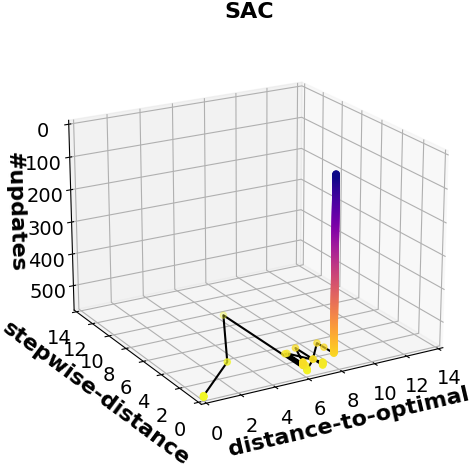} &   \includegraphics[width=0.18\linewidth]{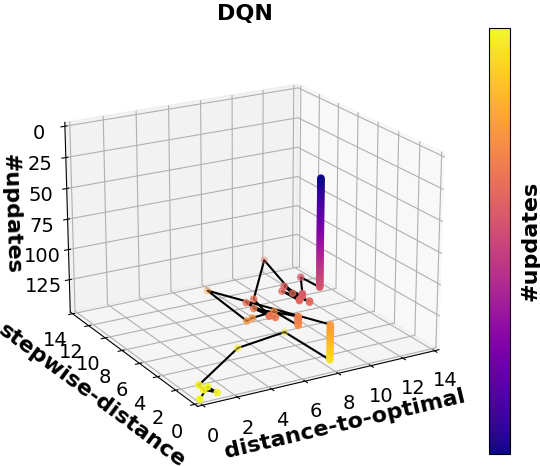} \\
     \includegraphics[width=0.16\linewidth]{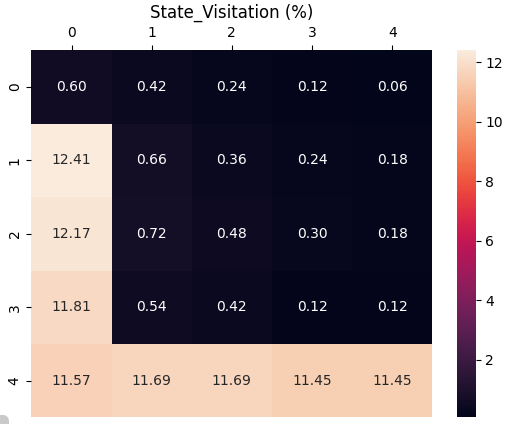} &   \includegraphics[width=0.16\linewidth]{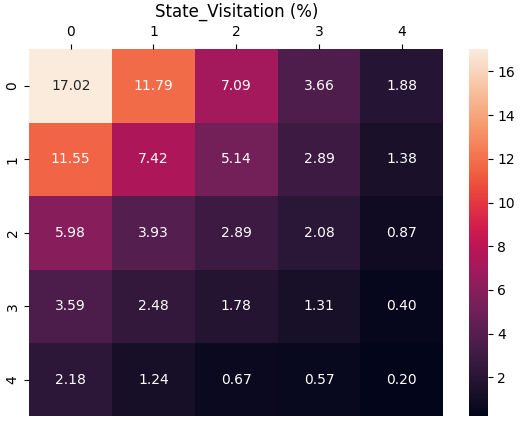} &   \includegraphics[width=0.16\linewidth]{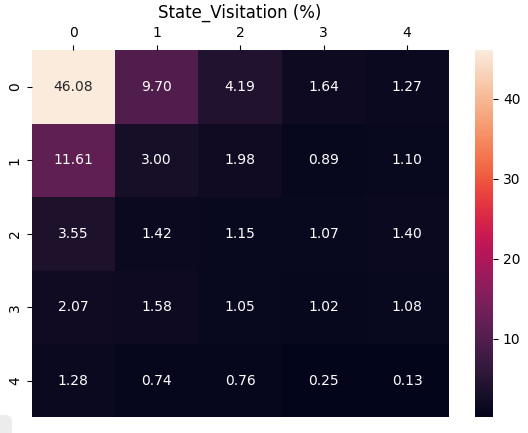} &  \includegraphics[width=0.16\linewidth]{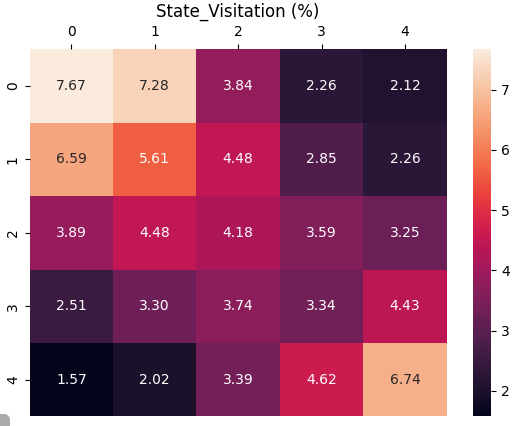} &  \includegraphics[width=0.16\linewidth]{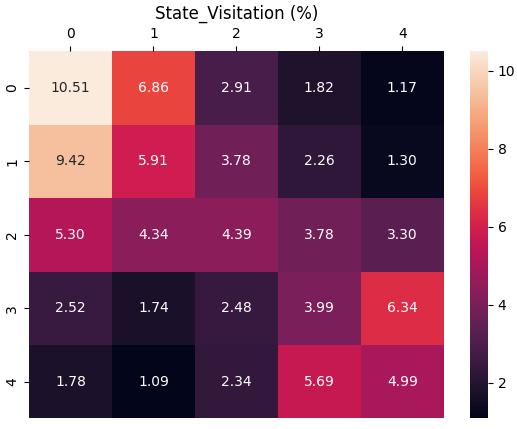} &  \includegraphics[width=0.16\linewidth]{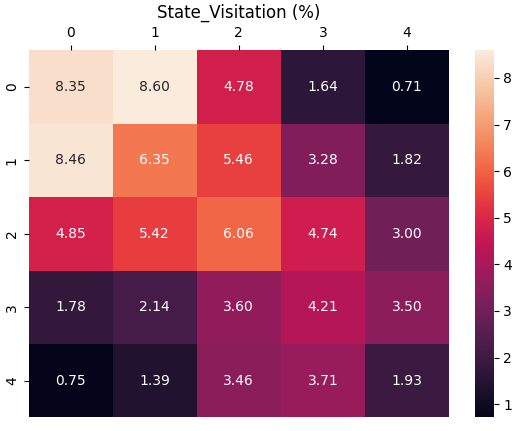} 
\end{tabular}
\caption{The top row showcases 3D scatter plots of \textit{distance-to-optimal} (x-axis) and \textit{stepwise-distance} (y-axis) across number of updates (z-axis), illustrating policy evolution in the occupancy measure space for RL algorithms: $\epsilon($=0)-greedy and $\epsilon($=1)-greedy Q-learning, UCRL2, PSRL, SAC, and DQN (left to right). The bottom row depicts the corresponding state visitation frequencies over the full training. The problem setting is deterministic with dense-rewards and 15 maximum number of steps per episode. (Larger versions of these plots are in Appendix~\ref{More Results})}\label{policy_evolution}
\vspace*{-1em}
\end{figure}

\vspace*{-.5em}
\underline{\textsc{(I) Discrete MDP.}}
We conducted an experiment with the following RL algorithms: 1) Tabular Q-learning with a) $\epsilon$-greedy ($\epsilon$ = 0) and b) $\epsilon$-greedy ($\epsilon$ = 1) strategies; 2) UCRL2 \citep{Jaksch10}; 3) PSRL \citep{Osband13}; 4) SAC \citep{Haarnoja18,Christodoulou19}; and 5) DQN \citep{Mnih13} with $\epsilon$-decay. The experiment was to solve the simple 5x5 gridworld with dense rewards, starting from top-left (0,0) reaching bottom-right (4,4). The aim of the exercise was to visualize the policy evolution of the RL algorithms in the occupancy measure space. In Figure~\ref{policy_evolution}, we observe behaviours of these RL algorithms in both the occupancy measure space and state space. 

\textbf{Q-learning: $\epsilon$ = 1 vs $\epsilon$ = 0.}
As expected, the state visitations show $\epsilon$ = 0 (which updates the Q-table by always exploiting) to have preferred visits that form a specific path from the start to the goal-state. In contrast, $\epsilon$ = 1 (which updates the Q-table by always exploring) exhibits visitation frequencies that are similar at states equidistant from the start-state, gradually decreasing as the distance from the state-state increases. The policy evolution of $\epsilon$ = 0 is more erratic and less likely to be stuck in suboptima compared to $\epsilon$ = 1. This is because high action randomness causes the agent to select suboptimal actions, slowing the Q-table convergence and making the learning policy appear stuck until the best actions are discovered.

\textbf{UCRL2 vs PSRL.} The state visits of UCRL2 are more uniform, with exception of the start-state because the initial state distribution is 1 at state (0,0). This is consistent with literature since the algorithm selects exploratory state-action pairs more uniformly~\citep{Jaksch10}. On the other hand, PSRL selects actions according to the probability that they are optimal~\citep{Osband13} which is observed based on high visit frequencies along the diagonals. Both algorithms have dispersed and erratic policy transitions, revealing that they search more aggressively by traversing broader within the occupancy measure space. They do not show any sign of being stuck or settling early on any particular policy. The exploration processes depict no interest of settling until the agent gains a good understanding of the environment. In this manner, UCRL2 and PSRL expend notable effort to reach optimality, especially UCRL2. The PSRL traverses fewer points than UCRL2 since it converges quicker, as per literature~\citep{Osband13}, and hence expends lower effort than UCRL2 (as shown in Table ~\ref{tab:env1}). 

\begin{figure}[t!]
\centering\vspace*{-2em}
\setlength{\arrayrulewidth}{0.1mm}
\setlength{\tabcolsep}{0.0mm}
\begin{tabular}{ll}    
\includegraphics[width=0.3\linewidth]{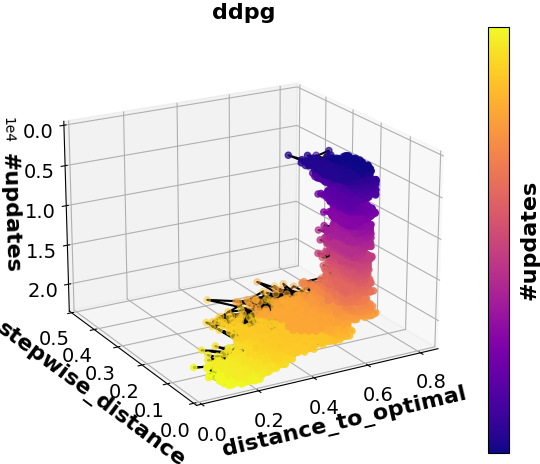} & \includegraphics[width=0.3\linewidth]{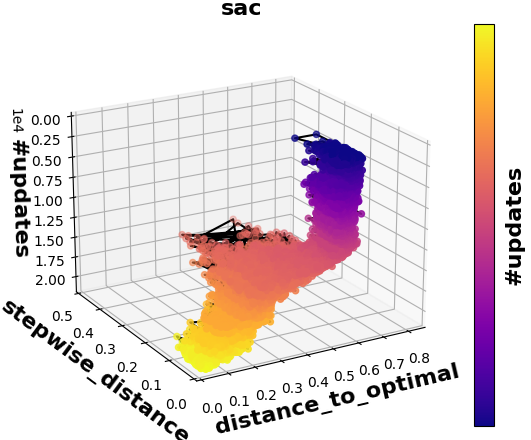} \\
\includegraphics[width=0.35 \linewidth]{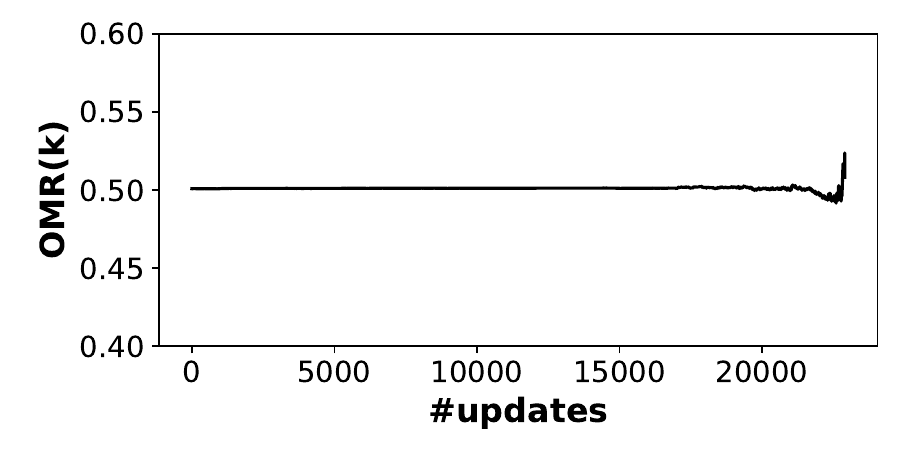} &
\includegraphics[width=0.35\linewidth]{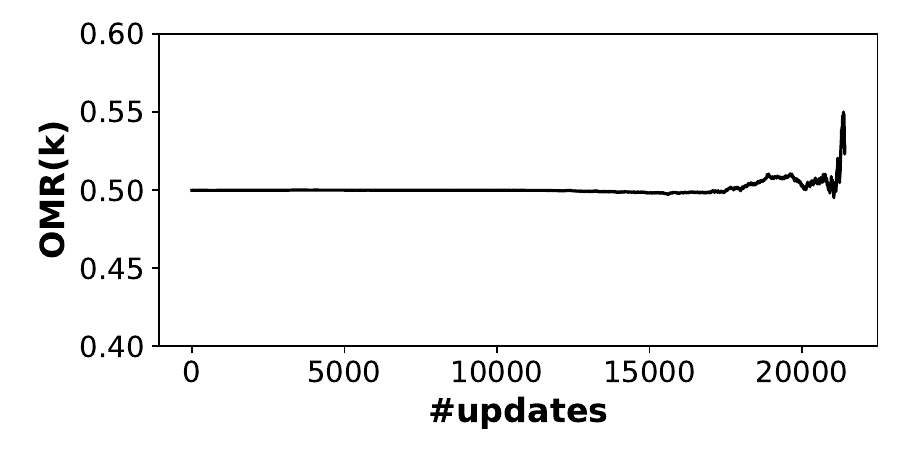} 
\end{tabular}\vspace*{-1.5em}
\caption{Top row: 3D scatter plots of \textit{distance-to-optimal} and \textit{stepwise-distance} over number of updates for algorithms DDPG and SAC. Bottom row: OMR($k$) versus \#update, $k$, for the corresponding algorithms.}\label{contin_policy_evolution}\vspace*{-1em}
\end{figure}
\setlength{\textfloatsep}{6pt}
\setlength{\arrayrulewidth}{0.5mm}
\setlength{\tabcolsep}{1mm}
\begin{figure}[t!]
\centering\vspace*{-2em}
\begin{minipage}{0.48\textwidth}
\begin{table}[H]
    \centering
    \resizebox{1.0\textwidth}{!}{
\begin{tabular}{lcccr}
\hline
Algo.& ESL & OMR & UC & SR\%\\
\hline
SAC   & 9.26$\pm$5.54 & 0.58$\pm$0.14 & 1480$\pm$670 & 100 \\
UCRL2 & {47.2}$\pm${8.20}$^{\star}$ & 0.49$\pm$0.04 & 60.7$\pm$11 & 100\\
PSRL  & 23.2$\pm$11.5 & 0.52$\pm$0.06 & \textbf{34.1}$\pm$\textbf{9.34} & 100 \\
DQN   & 12.4$\pm$7.13 & 0.54$\pm$0.11 & 226$\pm$93 & 98 \\
$\epsilon$(=1)-greedy& \textbf{6.27}$\pm$\textbf{2.22} & \textbf{0.61}$\pm$\textbf{0.09} & 672$\pm$385 & 100 \\
$\epsilon$(=0.9)-decay& 8.10$\pm$3.43 & 0.61$\pm$0.10 & 389$\pm$138 & 100 \\
$\epsilon$(=0)-greedy& 15.5$\pm$5.28 & 0.53$\pm$0.06 & 176$\pm$37.9 & 84 \\
\hline
\end{tabular}}
\caption{Evaluation of RL algorithms (over 40 runs) in the \textbf{deterministic, dense-rewards setting} for 5x5 gridworld, including Effort of Sequential Learning (ESL), Optimal Movement Ratio (OMR), number of updates to convergence (UC), and success rate (SR). Lowest ESL, highest OMR and lowest UC values are in \textbf{bold}, while the highest ESL value is starred ($\star$). 
}\label{tab:env1}
\end{table}
\end{minipage}\hfill
\begin{minipage}{0.5\textwidth}
\begin{table}[H]
    \centering
        \resizebox{1.0\textwidth}{!}{
\begin{tabular}{lcccr}
\hline
Algo. & ESL & OMR & UC & SR$\%$ \\
\hline
\multicolumn{5}{c}{\textbf{Deterministic, sparse}} \\
SAC   & \textbf{27.8}$\pm$\textbf{21.9} & \textbf{0.57}$\pm$\textbf{0.13} & 4885$\pm$3274 & 100 \\
UCRL2 & 73.3$\pm$0.0 & 0.45$\pm$0.0 & \textbf{93.0}$\pm$\textbf{0.0} & 100\\
PSRL  & 73.2$\pm$54.1 & 0.52$\pm$0.076 & 100$\pm$67.3 & 100 \\
DQN   & {137}$\pm${154}$^{\star}$ & 0.49$\pm$0.08 & 12703$\pm$4431 & 80 \\
\hline
\multicolumn{5}{c}{\textbf{Stochastic, dense} }\\
SAC   & {445}$\pm${245} & 0.501$\pm$0.004 & 2963$\pm$2043 & \textbf{92} \\
UCRL2 & 198$\pm$121 & 0.502$\pm$0.027 & 268$\pm$155 & 32\\
PSRL  & \textbf{55.4}$\pm$\textbf{33.6} & \textbf{0.52}$\pm$\textbf{0.04} & \textbf{76.1}$\pm$\textbf{50.6} & \textbf{92} \\
DQN   & 458$\pm$311$^{\star}$ & 0.502$\pm$0.01 & 1651$\pm$1077 & 24 \\
\hline
\end{tabular}}
\caption{Evaluation of RL algorithms (over 40 runs) in the \textbf{deterministic, sparse-rewards} and \textbf{stochastic, dense-rewards} settings for 5x5 gridworld. Lowest ESL, highest OMR and lowest UC values are in \textbf{bold}. The highest ESLs are starred.}\label{tab:env23}
\end{table}
\end{minipage}
\end{figure}

\textbf{SAC vs DQN.} The state visits of both the algorithms appear to be similar, however SAC manages to have higher visitation frequencies on the edges than DQN. Both algorithms start off without updating policies until the experience buffer is minimally filled. However, due to $\epsilon$ decay in DQN, it quickly converges since in this task acting more greedily is an advantage. SAC tends to stay at certain policies longer due to the slow ``soft updates" of its target networks. However, once it begins moving it reaches the optimal fast.  


In Figure~\ref{policy_evolution}, we see that UCRL2 explored the policy space more aggressively than the rest, which is corroborated by the highest ESL value in Table~\ref{tab:env1}. In Figure~\ref{policy_evolution}, we also see that the most efficient search processes (in terms of closely mimicking an oracle) in this setting, have minimal changes of the policy and more proportion of updates toward optimal, e.g. $\epsilon$-greedy, DQN and SAC, as corroborated by their ESL and OMR values in Table~\ref{tab:env1}. 

\underline{\textsc{(II) Continuous MDP.}} In this experiment, we use RL algorithms, DDPG~\citep{Lillicrap16} and SAC~\citep{Haarnoja18}, to solve the Mountain Car. The experiment aims to visualize the policy evolutions of these two RL algorithms in the occupancy measure space (Figure~\ref{contin_policy_evolution}).  

\textbf{DDPG vs SAC.} Both algorithms exhibit short-distances $(<1)$ between policy updates. The algorithms depict no sign of being stuck or settling early on any particular policy, which shows their continuously exploratory nature. While they begin with maintaining almost constant mean \textit{distances-to-optimal} and \textit{stepwise-distances}, SAC drops its mean \textit{distance-to-optimal} earlier than DDPG. Thus, SAC exhibits a lower ESL value than DDPG, which quantifies SAC's higher efficiency (Table \ref{tab:continuous}).

\begin{wraptable}{r}{0.4\textwidth}
    \centering\vspace*{-1em}
    \resizebox{0.4\textwidth}{!}{
\begin{tabular}{lcccr}
\hline
Algo.& ESL & OMR & UC & SR\%\\
\hline
DDPG   & 1881$\pm$500 & 0.501 & 23500 & 100 \\
SAC & 1619$\pm$189 & 0.5 & 22700 & 100\\
\hline
\end{tabular}}
\caption{Evaluation of RL algorithms in the Mountain Car continuous MDP (over 5 runs). The variances for OMR and UC are negligible. }\label{tab:continuous}\vspace*{-1em}
\end{wraptable}


Furthermore, we illustrate how OMR changes with update number $k$ in Figure~\ref{contin_policy_evolution}. OMR$(k)$ 
is defined as the OMR starting with the $k$th policy as the initial policy, whereas only OMR is from the $0$th policy. Details of computing OMR$(k)$ are provided in Appendix~\ref{temporal_OMR}. For both the algorithms, we observe that OMR$(k)$ stays almost constant at chance level ($\sim0.5$) in the beginning, and then, sharply increases towards the last updates. This observation indicates that the policy updates are initially oblivious to policy improvement (in terms of value functions) and purely exploratory, while they align towards the optimal policy just before convergence. The efficiency of the algorithm depends on how early this transition happens. For example, it starts earlier for SAC than DDPG rendering the first one more efficient.


\vspace*{-1em}\subsection{Comparison of ESL and OMR across RL Algorithms and Environments}\label{comparison_algos}\vspace*{-.5em}
Tables~\ref{tab:env1}-\ref{tab:continuous} showcase the evaluation of RL algorithms in various settings. 

\textbf{Dense Rewards.} We observe, in Table~\ref{tab:env1}, that PSRL took the lowest number of updates to reach the optimal policy in contrast with SAC. As explained in the previous section, SAC's "soft updates" of the target network slow down convergence. Despite the number of updates, SAC, DQN and Q-learning algorithms are meandering less as they move towards the optimal policies. The OMR of 0.61 in $\epsilon($=1)-greedy indicate that about 61\% of the algorithm's learning effort contributed to reducing the distance to the optimal policy. Generally, OMR values above 0.5 show that the algorithm spends more of its effort in reducing our regret analogue in the occupancy measure space.

\begin{wrapfigure}{r}{0.4\textwidth}
\centering\vspace*{-1em}
\centering
\includegraphics[width=0.4\textwidth]{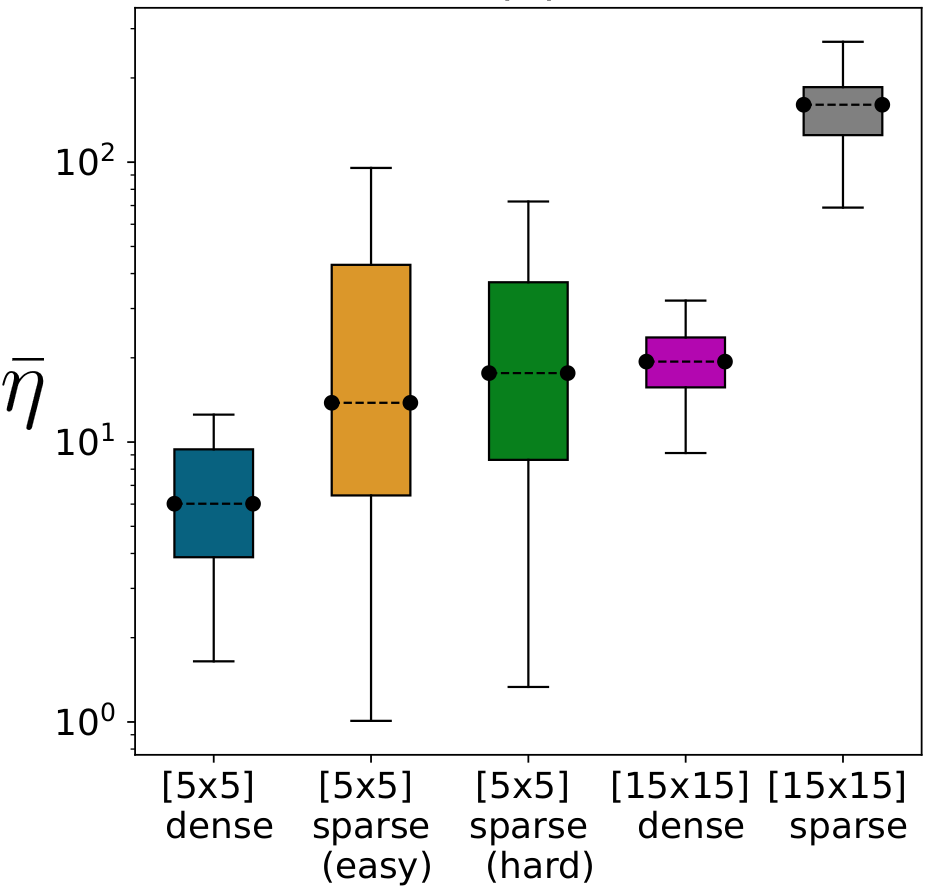}\vspace*{-.5em}
\caption{Q-learning with $\epsilon$-greedy ($\epsilon$ = 0.9 decaying, averaged over 40 runs) across deterministic 2D-Gridworld (5x5 and 15x15) tasks. The 1st and 4th (from left to right) have dense rewards, while the rest have sparse rewards (details in Appendix~\ref{Environment Description}).}\label{envir}\vspace*{-3em}
\end{wrapfigure}

\textbf{Sparse Rewards.} In the sparse rewards setting (Table~\ref{tab:env23}), SAC is more efficient with lowest ESL and highest OMR. UCRL2 has the lowest number of updates. DQN has the worst overall performance depicted by low success rate, high number of updates, ESL and OMR. From the dense-rewards to the sparse-rewards settings, we notice a general increase in ESL values and decrease in OMR across algorithms. This makes sense since the task hardness increases.  

\textbf{Stochastic Transitions.} In the stochastic setting (Table~\ref{tab:env23}), PSRL leads in all metrics, along with a high success rate. In all the settings, lower ESL values correlate with higher OMR values, as a more efficient agent travels more directly towards the optimal policy. The number of policy updates does not reveal much about the policy's evolution, whereas ESL indicates whether the agent is mostly meandering or moving directly towards optimality. Once again, we notice a significant increase in ESL values when shifting from deterministic to stochastic transitions, indicating an increase in the effort of the exploratory process. 



\vspace*{-1 em}
\subsection{ESL Increases with Task Difficulty}\label{tasks_section}\vspace*{-.5em}
Figure \ref{envir} illustrates the ESL values for Q-learning with $\epsilon$-decay strategy (for $\epsilon$ = 0.9) across tasks with varying hardness. 
These tasks are deterministic 2D-Gridworld of sizes 5x5 and 15x15 matched with either dense or sparse rewards (as specified in Appendix~\ref{Environment Description}).  We chose to assess the $\epsilon$-decay Q-learning algorithm because it is simple and yet completes all these tasks. We observe that the ESL is lowest for \textit{[5x5] dense} (5x5 grid, dense rewards) and highest for \textit{[15x15] sparse} (15x15 grid, sparse rewards) as anticipated.  The results demonstrate that ESL scales proportionally with task difficulty, matching expectations that more difficult tasks demand greater effort of the exploratory process.

\vspace*{-.5em}\section{Related Works}\vspace*{-.5em}
Several prior works have utilized various components leveraged in our work, namely Wasserstein distance, occupancy measures, and the trajectory of RL on a manifold, but for different purposes. 
Here, we summarise them and elucidate the connections.

In supervised learning, \citet{Alvarez20} proposed an optimal transport approach, namely Optimal Transport Dataset Distance (OTDD), to quantify the transferability between two supervised learning tasks by computing the similarity (aka distance) between the task datasets. Here, we conceptualise and define the effort of learning for RL, as a sequence of such supervised learning tasks. We observe that \textit{the total effort of sequential learning can be computed as the sum of OTDD distances between consecutive occupancy measures}. Recently, \citet{Zhu24} have developed generalized occupancy models by defining cumulative features that are transferable across tasks. In future, one can generalize our indices for the cumulative features constructed from some invertible functions of the step-wise occupancy measures. 

Optimal transport-based approaches are also explored in RL literature. These works broadly belong to two families. First line of works uses Wasserstein distance over a posterior distribution of Q-values~\citep{Metelli19,Likmeta23} or return distributions~\citep{Sun22} to quantify uncertainty, and then to use this Wasserstein distance as a loss to learn better models of the posterior distribution of Q-values or return distributions, respectively. 
The second line of works uses Wasserstein distance between a feasible family of MDPs as an additional robustness constraint to design robust RL algorithms~\citep{Abdullah19,Derman20,Hou20}. Here, \textit{we bring a novel concept of using Wasserstein distance between occupancy measures to understand the exploratory dynamics}. Incorporating this insight into better algorithm design would be an interesting future work. 
Recently, \citet{Calo24} relate Wasserstein distance between reward-labelled Markov chains to bisimulation metrics which abstract state spaces. In the same spirit, we could use reward as the cost-function in computing our nested Wasserstein distance (OTDD) to obtain a reward- or value-aware OTDD to define broader bisimulation metrics with abstract state-action spaces, instead of just state spaces. 

As a parallel approach to optimal transport, the information geometries of the trajectory of an RL algorithm under different settings are studied. These approaches use mutual information as a metric instead of Wasserstein distance. \citet{Basu20} study the information geometry of Bayesian multi-armed bandit algorithms. They consider a bandit algorithm as a trajectory on a belief-reward manifold, and propose a geometric approach to design a near-optimal Bayesian bandit algorithm. \cite{Eysenbach21,Laskin22} study information geometry of unsupervised RL and propose mutual information maximization schemes over a set of tasks and their marginal state distributions. \cite{Yang24} extend this approach with Wasserstein distance and demonstrate benefits of using Wasserstein distance than mutual information. \textit{We use Wasserstein distance as a natural metric in occupancy manifold that also allows comparison of hardness of different tasks.} It would be interesting to extend our framework to understand the dynamics of unsupervised RL algorithms.

\vspace*{-.5em}\section{Discussion and Future Works}\vspace*{-.5em}
Our work introduces methods to theoretically and quantitatively understand and compare the learning strategies of different RL algorithms. Since learning in a typical RL algorithm happens through a sequence of policy updates, we propose to understand the learning process by visualizing and analysing the path traversed by an RL algorithm in the space of occupancy measures corresponding to this sequence. 




We show the usefulness of this approach by conducting experiments on various environments. Our results show that the indices ESL and OMR provide insight into the agent's policy evolution, revealing whether it is steadily approaching the optimal policy or mostly meandering. Additionally, this allows us to understand how the learning process of the same algorithm changes with different rewards and transitions structures, and task hardness. A key limitation of our indices is that they are based on assumption that the final policy reached at the end of training is an optimal one, though we could still derive some benefit from our approach even if not. In the future, it would be interesting to use this approach to benchmark and compare the learning dynamics of different RL algorithms on further environments. In addition, it would be useful to study whether the occupancy measures trajectory of an algorithm provides insights to improve its exploratory process.

\subsubsection*{Acknowledgments}
R. Nkhumise thanks Pawel Pukowski for helpful discussions on initial theoretical aspects. R. Nkhumise was supported by the EPSRC Doctoral Training Partnership (DTP) - Early Career Researcher funding awarded to A. Gilra. A. Gilra acknowledges the CHIST-ERA grant for the Causal Explanations in Reinforcement Learning (CausalXRL) project (CHIST-ERA-19-XAI-002), by the Engineering and Physical Sciences Research Council, United Kingdom (grant reference EP/V055720/1) for supporting the work. D. Basu acknowledges the CHIST-ERA grant for the CausalXRL project (CHIST-ERA-19-XAI-002) by L'Agence Nationale de la Recherche, France (grant reference ANR-21-CHR4-0007), the ANR JCJC for the REPUBLIC project (ANR-22-CE23-0003-01), and the PEPR project FOUNDRY (ANR23-PEIA-0003) for supporting the work. We thank Eleni Vasilaki and Philippe Preux for their support.

\bibliography{iclr2025_conference}
\bibliographystyle{iclr2025_conference}

\appendix
\clearpage
\onecolumn

\section{Theoretical Analysis}
\subsection{MDP with Lipschitz Rewards}
\label{Appendix Lipschitz MDP}
Given two metric spaces $(\mathcal{X}, d_{\mathcal{X}})$ and $(\mathcal{Y}, d_{\mathcal{Y}})$, a function $f: \mathcal{X} \to \mathcal{Y}$ is called 1-Lipschitz continuous if \citep{Villani09}: 
\begin{equation}\label{lipschitz-1}
\begin{aligned}
    d_{Y}(f(x),f(x')) \leq d_{X}(x,x') \text{, } \forall (x,x') \in X 
\end{aligned} 
\end{equation}

This implies that the Lipschitz semi-norm over the function space $\mathcal{F}(X,Y)$, defined as 
\begin{equation}\label{lipschitz-semi_norm}
\begin{aligned}
    \Vert f \Vert_{L}  = \sup_{x \neq x'} \left\{ \frac{d_{Y}(f(x),f(x'))}{d_{X}(x,x')}  \mid \forall (x,x') \in \mathcal{X} \right\} ,
\end{aligned} 
\end{equation}
is $ \leq 1$. When $(\mathcal{X}, d_{\mathcal{X}})$ is a Polish space and $\mu,\nu \in \mathcal{P}(\mathcal{X})$, the \textbf{Kantorovich-Rubinstein} formula states that  \citep{Villani09}: 
\begin{equation}\label{kantorovich-Rubinstein}
\begin{aligned}
    \mathcal{W}_{1}(\mu,\nu) 
    &= \sup_{\Vert f \Vert_{L} \leq 1} \left\{ \int_{\mathcal{X}} f \,d\mu - \int_{\mathcal{X}} f \,d\nu \right\} \\
    &= \sup_{\Vert f \Vert_{L} \leq 1} \left\{ \mathbb{E}_{\mu} \left[ f(X) \right] - \mathbb{E}_{\nu} \left[ f(X) \right] \right\},
\end{aligned}
\end{equation}
where $W_1(\mu,\nu)$ is the 1-Wasserstein distance between $\mu$ and $\nu$ with $f$ as the cost function.

Note that when $\Vert f \Vert_{L} \leq L_{\mathcal{R}}$ for any $L_{\mathcal{R}} > 0$, then function $f$ is called $L_{\mathcal{R}}$-Lipschitz continuous, and \Eqref{kantorovich-Rubinstein} becomes~\citep{Gelada19}, 
\begin{equation}\label{kantorovich-Rubinstein_2}
\begin{aligned}
    \mathcal{W}_{1}(\mu,\nu) 
    &= \frac{1}{L_{\mathcal{R}}} \sup_{\Vert f \Vert_{L} \leq L_{\mathcal{R}}} \left\{ \mathbb{E}_{\mu} \left[ f(X) \right] - \mathbb{E}_{\nu} \left[ f(X) \right] \right\}.
\end{aligned}
\end{equation}

Now, we consider $\mathcal{X}=\mathcal{S} \times \mathcal{A}$, i.e. the state-action space, $\mathcal{Y}=\mathbb{R}$, i.e. the real line, and the function $f$ to be the reward function $\mathcal{\bar{R}}$.
Then, we can call the reward function $\mathcal{\bar{R}}$ to be $L_{\mathcal{R}}$-Lipschitz if
\[|\mathcal{\bar{R}}(s,a) - \mathcal{\bar{R}}(s',a')| \leq L_{\mathcal{R}}d_{\mathcal{S}\mathcal{A}}((s,a),(s',a'))\]
for all $s, s' \in \mathcal{S}$, and $a, a' \in \mathcal{A}$, 
and $d_{\mathcal{S}\mathcal{A}}((s,a),(s',a')) = d_{\mathcal{S}}((s,s')) + d_{\mathcal{A}}((a,a'))$ being the metric on the state-action space $\mathcal{S} \times \mathcal{A}$. 
If the reward function $\mathcal{\bar{R}}$ of an MDP is $L_{\mathcal{R}}$-Lipschitz, we refer it as an MDP with Lipschitz rewards.

\subsection{Performance Difference and Occupancy Measures}
\label{Appendix  Performance Difference}
We know that
\begin{equation} \label{objective_occupancy}
J^{\pi}_{\mu} = \frac{1}{\rho} \mathbb{E}_{(s,a) \sim v_{\pi}} \left[ \mathcal{\bar{R}}(s,a) \right]\,.
\end{equation}
Using \Eqref{objective_occupancy}, we write for two policies $\pi$ and $\pi'$, with $\mu(s)$ as the initial state distribution,
\begin{equation}\label{objective_occupance_proof_1}
\begin{aligned}
    \left| J^{\pi}_{\mu} - J^{\pi'}_{\mu} \right| 
    &= \frac{1}{\rho} \left| \mathbb{E}_{(s,a) \sim v_{\pi}} \left[ \mathcal{\bar{R}}(s,a) \right] - \mathbb{E}_{(s,a) \sim v_{\pi'}} \left[ \mathcal{\bar{R}}(s,a) \right] \right|
\end{aligned}
\end{equation}

Given an MDP with $L_{\mathcal{R}}$-Lipschitz rewards, the \textbf{Kantorovich-Rubinstein} formula dictates that \citep{Gelada19}:  
\begin{equation}\label{objective_occupance_proof_2}
\begin{aligned}
    \sup_{\Vert \bar{R} \Vert_{L} \leq  L_{\mathcal{R}}} \left|
    \mathbb{E}_{(s,a) \sim v_{\pi}} \left[ \mathcal{\bar{R}}(s,a) \right] - \mathbb{E}_{(s,a) \sim v_{\pi'}} \left[ \mathcal{\bar{R}}(s,a) \right] \right|
    &= L_{\mathcal{R}} \mathcal{W}_{1}(v_{\pi},v_{\pi'})
\end{aligned}
\end{equation}

By dividing both sides of \Eqref{objective_occupance_proof_2} by $\rho$, and due to an upper bound by the supremum, this inequality follows:
\begin{equation}\label{objective_occupance_proof_3}
\begin{aligned}
    &\left| J^{\pi}_{\mu} - J^{\pi'}_{\mu} \right| \leq\frac{L_{\mathcal{R}}}{\rho} \mathcal{W}_{1}(v_{\pi},v_{\pi'})
\end{aligned}
\end{equation}

\subsection{Proof of proposition 1}
\label{Appendix prop 1}
The Linear Programming formulation for solving MDPs, assuming discrete state and action spaces, is \citep{Puterman94}: 

\begin{equation}\label{lp}
\begin{aligned}
    \max_{v_{\pi}} &\sum_{s,a}r(s,a)v_{\pi}(s,a) \\
     \text{subject to} 
    &\sum_{a}v_{\pi}(s,a) = p_{0}(s) + \gamma \sum_{s',a}T(s \mid s',a)v_{\pi}(s',a) \\
    &v_{\pi}(s,a) \ge 0~~~~\forall (s,a) \in \mathcal{S} \times \mathcal{A}\,,
\end{aligned} 
\end{equation}
where $p_{0}(s)$ is the initial state distribution and $T(s \mid s',a)$ is the transition probability. The constraints of this optimization problem are often referred to as \textit{Bellman Flow Constraint}. 

A stationary policy $\pi$ has a corresponding occupancy measure $v_{\pi}(s,a)$ that satisfies the Bellman flow constraint \citep{Syed08}, and hence $\pi$ and $v_{\pi}(s,a)$ share a bijective relationship \citep{Syed08,Givchi21},

\begin{equation} \label{pol_occ}
    \pi(a \mid s) = \frac{v_{\pi}(s,a)}{ u_{\pi}(s)}
\end{equation}
with 
\begin{equation} \label{state_measures}
    u_{\pi}(s) = \sum_{a'}v_{\pi}(s,a') = p_{0}(s) + \gamma \sum_{s',a'}T(s \mid s',a')v_{\pi}(s',a')
\end{equation}

By rearranging \Eqref{pol_occ} to 
\begin{equation} \label{occ_pol}
    v_{\pi}(s,a) = \pi(a \mid s)u_{\pi}(s) 
\end{equation}
and substituting \Eqref{occ_pol} into \Eqref{state_measures}, we can rewrite \Eqref{state_measures} as (defining $\mathcal{P}^\pi \triangleq \sum_{a}T(s \mid s',a)\pi(a \mid s')$),
\begin{equation} \label{state_measures_2}
\begin{aligned}
p_{0}(s) &= u_{\pi}(s) - \gamma \sum_{s',a}T(s \mid s',a)\pi(a \mid s')u_{\pi}(s') \\
&\triangleq u_{\pi}(s) - \gamma \sum_{s'}\mathcal{P}^{\pi}(s \mid s')u_{\pi}(s') \\
\end{aligned} 
\end{equation}
which in matrix form is
\begin{equation} \label{matrix_form_state_measure}
\begin{aligned}
\mathbf{p}_{0} &= \mathbf{u}_{\pi} - \gamma \mathbf{P^{\pi}}\mathbf{u}_{\pi} \\
&= \left( \mathbb{I} - \gamma \mathbf{P^{\pi}} \right)\mathbf{u}_{\pi} \,,
\end{aligned}
\end{equation}
where $\mathbf{p}_{0}, \mathbf{u}_{\pi} \in \mathbb{R}^{|\mathcal{S}|}$ are column vectors and $\mathbf{P^{\pi}} \in \mathbb{R}^{|\mathcal{S}| \times |\mathcal{S}| }$ are matrices. Solving for $\mathbf{u}_{\pi}$, we get
\begin{equation} \label{matrix_u}
    \mathbf{u}_{\pi} = \left( \mathbb{I} - \gamma \mathbf{P^{\pi}} \right)^{-1}\mathbf{p}_{0}
\end{equation}

The inverse matrix $ \left( \mathbb{I} - \gamma \mathbf{P^{\pi}} \right)^{-1}$ exists because for $\gamma < 1$, $ \left( \mathbb{I} - \gamma \mathbf{P^{\pi}} \right)$ is  a strictly diagonally dominant matrix~\citep{Syed08}. Thus, $\left( \mathbb{I} - \gamma \mathbf{P^{\pi}} \right)^{-1} = \sum_{t=0}^{\infty}( \gamma \mathbf{P^{\pi}})^{t}$, where $\sum_{t=0}^{\infty}( \gamma \mathbf{P^{\pi}})^{t}$ forms a valid \textit{Neumann series} \citep{Ward21}. We let $\mathbf{A}^{\pi} = \sum_{t=0}^{\infty}( \gamma \mathbf{P^{\pi}})^{t}$, so \Eqref{matrix_form_state_measure} can be written as $\mathbf{u}_{\pi} = \mathbf{A}^{\pi}\mathbf{p}_{0}$. We can therefore express \Eqref{occ_pol} in matrix form as:

\begin{equation}\label{occ_pol_matrix}
\begin{aligned}
    \mathbf{v}_{\pi}  &= \mathbf{\Pi} \odot \left( \mathbf{u}_{\pi}^{T} \otimes  \mathbf{1} \right)^{T} \\
    &= \mathbf{\Pi} \odot \left( \mathbf{p}_{0}^{T}(\mathbf{A}^{\pi})^{T} \otimes  \mathbf{1} \right)^{T}\,,
\end{aligned}
\end{equation}
where $\mathbf{\Pi}, \mathbf{v}_{\pi} \in \mathbb{R}^{|\mathcal{S}| \times |\mathcal{A}|}$,  $\mathbf{1} \in \mathbb{R}^{|\mathcal{A}|}$ is a column vector of ones, $\otimes$ presents the Kronecker product, and $\odot$ denotes the Hadamard product.

If we consider the case of a parameterized policy $\mathbf{\Pi} ({\theta})$, then the derivative of $\mathbf{v}_{\pi}$ with respect to $\theta$ is
\begin{equation} \label{diff_occ_pol_2}
\begin{aligned}
    \nabla_{\theta} \mathbf{v}_{\pi} =& \nabla_{\theta} \left[ \mathbf{\Pi} \odot \left( \mathbf{p}_{0}^{T}(\mathbf{A}^{\pi})^{T} \otimes  \mathbf{1} \right)^{T} \right] \\
    =& \nabla_{\theta} \mathbf{\Pi} \odot \left( \mathbf{p}_{0}^{T}(\mathbf{A}^{\pi})^{T} \otimes  \mathbf{1} \right)^{T} + \mathbf{\Pi} \odot \nabla_{\theta}\left( \mathbf{p}_{0}^{T}(\mathbf{A}^{\pi})^{T} \otimes  \mathbf{1} \right)^{T}  \\
    =& \nabla_{\theta} \mathbf{\Pi} \odot \left( \mathbf{p}_{0}^{T}(\mathbf{A}^{\pi})^{T} \otimes  \mathbf{1} \right)^{T}  + \mathbf{\Pi} \odot \left( \mathbf{p}_{0}^{T}(\nabla_{\theta}\mathbf{A}^{\pi})^{T} \otimes  \mathbf{1} \right)^{T}  
\end{aligned}
\end{equation}

The first term in \Eqref{diff_occ_pol_2} is differentiable since the policy is parameterized by $\theta$. We expand $\nabla_{\theta}\mathbf{A}^{\pi}$ as follows:
\begin{equation} \label{a_expand}
\begin{aligned}
    \nabla_{\theta}\mathbf{A}^{\pi} &= \sum_{t=0}^{\infty} t (\gamma \mathbf{P}^\pi)^{t-1}\gamma \nabla_\theta \mathbf{P}^\pi \\
    &\equiv \sum_{t=0}^{\infty} t (\gamma \mathbf{P}^\pi)^{t-1} \gamma\nabla_\theta \left[ \sum_a T(s|s',a)\pi(a|s') \right] \\
    &= \sum_{t=0}^{\infty} t (\gamma \mathbf{P}^\pi)^{t-1} \gamma \left[\sum_a T(s|s',a)\nabla_\theta\pi(a|s') \right] \\
    &= \sum_{t=0}^{\infty} t (\gamma \mathbf{P}^\pi)^t ({\mathbf{P}^\pi})^{-1}\left[ \sum_a T(s|s',a)\nabla_\theta\pi(a|s') \right]
\end{aligned}
\end{equation}

If $({\mathbf{P}^\pi})^{-1}$ exists, then $\nabla_{\theta}\mathbf{A}^{\pi}$ is differentiable, and consequently so is $\nabla_{\theta} \mathbf{v}_{\pi}$, based on \Eqref{diff_occ_pol_2} and \Eqref{a_expand}. Proceeding similarly, given the same conditions, we see that all higher derivatives of $v_\pi$ also exist with respect to $\theta$. Thus, the space of parametrized occupancy measures $v_\pi$ forms a differentiable manifold.

\subsection{Proof of proposition 2}
\label{Appendix prop 2}

Regret is a common metric for evaluating agents, that measures the total loss an agent incurs over policy updates by using its policy in lieu of the optimal one, defined as~\citep{Osband13},
\begin{equation}\label{cum_regret}
\begin{aligned}
    \text{Regret} = \mathbb{E}_{s \sim \mu}\left[  \sum_{k} ( V^{*}(s) - V_{\pi_{k}}(s))\right] 
\end{aligned} 
\end{equation}
where $V^{*} = V_{\pi^{*}}$ is the value function of the optimal policy $\pi^*$ while $V_{\pi_{k}}(s)$ is the value function of policy $\pi_{k}$, and $\mu$ is the initial state distribution.  

Since ${J^{\pi}_{\mu} = \mathbb{E}_{s \sim \mu} [V_{\pi}(s)]}$, we can conclude from \Eqref{cum_regret} that
\begin{equation}\label{cum_regret_2}
\begin{aligned}
    \text{Regret} &= \mathbb{E}_{s \sim \mu}\left[  \sum_{k} ( V^{*}(s) - V_{\pi_{k}}(s))\right]  \\
    &= \sum_{k} \left[   \mathbb{E}_{s \sim \mu} ( V^{*}(s) - V_{\pi_{k}}(s))\right] \\
    &= \sum_{k} \left(   J^{\pi^*}_{\mu} - J^{\pi_k}_{\mu} \right) \\
    &= \sum_{k} \left|   J^{\pi^*}_{\mu} - J^{\pi_k}_{\mu} \right| \\
    &\leq \sum_{k} \frac{L_{\mathcal{R}}}{\rho} \mathcal{W}_{1}(v_{\pi^*},v_{\pi_k})
\end{aligned} 
\end{equation}

The last inequality is due to \Eqref{objective_occupance_proof_3}.

\subsection{Proof of proposition 3}
\label{Appendix prop 3}

Let us begin the proof by defining the visitation probability at any step $h \in [H]$ in an episode, following policy $\pi(a|s)$. Specifically,
\begin{equation}
    q^h_{\pi}(s,a) \triangleq \prob(s_h=s, a_h=a)~~\forall h \in[H]~~\text{ and }~~q^h_{\pi}(s,a) \triangleq 0~~\forall h \in \mathbb{N} \wedge h > H\,.
\end{equation}
Thus, we rewrite \Eqref{episodic_occupancy} $v^H_{\pi}(s,a) = \frac{1}{H} \sum_{h=1}^H q^h_{\pi}(s,a)$.

Then, following~\citep{Kalagarla21}, we can write the Linear Programming formulation for solving episodic MDP $\mathbb{M}^H$ as
\begin{equation}\label{lp_finite}
\begin{aligned}
    \max_{\{q^h_{\pi}\}_{h=1}^H} &\sum_{h,s,a}r(s,a)q^h_{\pi}(s,a) \\
     \text{subject to} 
    &\sum_{a}q^{h}_{\pi}(s,a) = \sum_{s',a}T(s \mid s',a)q^{h-1}_{\pi}(s',a)~~~\forall h \in [H] \wedge h >1\,,\\
    &q^1_{\pi}(s,a) = \pi(a|s)\mu(s)\,,\\
    &q^h_{\pi}(s,a) \ge 0 ~~~~~~~~~~\forall h \in [H], (s,a) \in \mathcal{S}\times\mathcal{A}\,,
\end{aligned} 
\end{equation}
where $\mu(s)$ is the initial state distribution and $T(s \mid s',a)$ is the transition probability. The constraints of this optimization problem are often referred to as \textit{Bellman Flow Constraints}. 

This implies that 
\begin{align}
    &\sum_{h=2}^{H+1}\sum_{a}q^{h}_{\pi}(s,a) = \sum_{h=2}^{H+1}\sum_{s',a}T(s \mid s',a)q^{h-1}_{\pi}(s',a) \notag\\
    \implies &\sum_a q^{1}_{\pi}(s,a) + \sum_{h=2}^{H+1}\sum_{a}q^{h}_{\pi}(s,a) = \sum_{h=2}^{H+1}\sum_{s',a}T(s \mid s',a)q^{h-1}_{\pi}(s',a) + \sum_a q^{1}_{\pi}(s,a)\notag\\
    \implies  &\sum_a \sum_{h=1}^{H+1} q^{h}_{\pi}(s,a) = \sum_{h=2}^{H+1}\sum_{s',a}T(s \mid s',a)q^{h-1}_{\pi}(s',a) + \sum_a q^{1}_{\pi}(s,a)\notag\\
    \implies  &H \sum_a v^{H}_{\pi}(s,a) = \sum_{s',a}T(s \mid s',a)(\sum_{h=2}^{H+1} q^{h-1}_{\pi}(s',a)) + \mu(s)\notag\\
    \implies  &H \sum_a v^{H}_{\pi}(s,a) = H \sum_{s',a}T(s \mid s',a)v^{H}_{\pi}(s',a) + \mu(s) \notag\\
    \implies  &\sum_a v^{H}_{\pi}(s,a) = \sum_{s',a}T(s \mid s',a)v^{H}_{\pi}(s',a) + \frac{1}{H} \mu(s) \notag\\
    \implies  &u^H_{\pi}(s) \triangleq \sum_a v^{H}_{\pi}(s,a) = \sum_{s',a}T(s \mid s',a)\pi(a|s') u^{H}_{\pi}(s') + \frac{1}{H} \mu(s) \,.
\end{align}
Now, we denote $\mathbf{u}^H_\pi$ and $\mathbf{\mu}$ as corresponding column vectors and $P^\pi(s,s') \triangleq\sum_{s',a}T(s \mid s',a) \pi(a|s')$ as matrix. Thus, we obtain

\begin{equation}  
\begin{aligned}
    (\mathbb{I} - P^{\pi}) \mathbf{u}^H_{\pi} &= \frac{1}{H} \mathbf{\mu} 
    \implies \mathbf{u}^H_{\pi} = \frac{1}{H} (\mathbb{I} - P^\pi)^{-1} \mathbf{\mu}\,.
\end{aligned}
\end{equation}

We can therefore express the finite horizon occupancy measure in matrix form as 
\begin{equation} \label{occ_pol_matrix}
\begin{aligned}
    \mathbf{v}^H_{\pi}  &= \mathbf{\Pi} \odot \left( (\mathbf{u}^H_{\pi})^T \otimes  \mathbf{1} \right)^T 
    = \mathbf{\Pi} \odot \left( \mu^{T}(\mathbf{A}^{\pi}_H)^{T} \otimes  \mathbf{1} \right)^T \\
\end{aligned}
\end{equation}
where $\mathbf{\Pi}, \mathbf{v}_{\pi} \in \mathbb{R}^{|\mathcal{S}| \times |\mathcal{A}|}$,  $\mathbf{1} \in \mathbb{R}^{|\mathcal{A}|}$ is a column vector of ones, $\otimes$ presents the Kronecker product, $\odot$ denotes the Hadamard product, and $\mathbf{A}^{\pi}_H \triangleq \frac{1}{H} (\mathbb{I} - P^\pi)^{-1}$.

If we consider the case of a parameterized policy $\mathbf{\Pi} ({\theta})$, the derivative of $\mathbf{v}_{\pi}$ with respect to $\theta$ is
\begin{equation} \label{diff_occ_pol_3}
\begin{aligned}
    \nabla_{\theta} \mathbf{v}_{\pi} =& \nabla_{\theta} \left[ \mathbf{\Pi} \odot \left( \mathbf{p}_{0}^{T}(\mathbf{A}^{\pi})^{T} \otimes  \mathbf{1} \right)^T \right] \\
    =& \nabla_{\theta} \mathbf{\Pi} \odot \left( \mathbf{p}_{0}^{T}(\mathbf{A}^{\pi})^{T} \otimes  \mathbf{1} \right)^T + \mathbf{\Pi} \odot \nabla_{\theta}\left( \mathbf{p}_{0}^{T}(\mathbf{A}^{\pi})^{T} \otimes  \mathbf{1} \right)^T  \\
    =& \nabla_{\theta} \mathbf{\Pi} \odot \left( \mathbf{p}_{0}^{T}(\mathbf{A}^{\pi})^{T} \otimes  \mathbf{1} \right)^T + \mathbf{\Pi} \odot \left( \mathbf{p}_{0}^{T}(\nabla_{\theta}\mathbf{A}^{\pi})^{T} \otimes  \mathbf{1} \right)^T \\
\end{aligned}
\end{equation}

The first term in \Eqref{diff_occ_pol_3} is differentiable since the policy is parameterized by $\theta$. We expand on $\nabla_{\theta}\mathbf{A}^{\pi}$ as follows:
\begin{equation} \label{a_expand_2}
\begin{aligned}
    H~\nabla_{\theta}\mathbf{A}^{\pi} &=  \nabla_{\theta} (\mathbb{I} - P^\pi)^{-1}\\
    &= \nabla_{\theta} \left(\sum_{i=0}^{\infty} (P^{\pi})^{i}\right)\\
    &=  \sum_{i=0}^{\infty} i (P^{\pi})^{i-1} \nabla_{\theta}P^{\pi}\\
    &=  \sum_{i=0}^{\infty} i (P^{\pi})^{i-1} T \nabla_{\theta}\Pi\,.
\end{aligned}
\end{equation}

If $(P^{\pi})^{-1}$ exists, then $\nabla_{\theta}\mathbf{A}^{\pi}$ is differentiable, and consequently so is $\nabla_{\theta} \mathbf{v}^H_{\pi}$. Proceeding similarly, given the same conditions, we see that all higher derivatives of $\mathbf{v}^H_\pi$ also exist with respect to $\theta$. Thus, the space of parametrized finite-horizon occupancy measures $v^H_\pi$ forms a differentiable manifold $\mathcal{M}^H$.

\subsection{Optimal Transport Dataset Distance (OTDD)}
\label{Appendix OTDD}
Suppose we have two datasets, each consisting of feature-label pairs, $\mathcal{D}_{A} = \{ (t_{A}^{i},u_{A}^{i}) \}_{i=1}^{m} \sim P_{A}(t,u)$ and $\mathcal{D}_{B} = \{ (t_{B}^{i},u_{B}^{i}) \}_{i=1}^{n} \sim P_{B}(t,u)$ with $t_A, t_B \in \mathcal{T}$ and $u_A, u_B \in  \mathcal{U_{A}}, \mathcal{U_{B}} $. These datasets can be used to create empirical distributions $\hat{P}_{A}(t,u)$ and $\hat{P}_{B}(t,u)$. OTDD is the p-Wasserstein distance between the datasets $\mathcal{D}_{A}$ and $\mathcal{D}_{B}$ - which is essentially the distance between their empirical distributions $\hat{P}_{A}$ and $\hat{P}_{B}$ - with the cost function defined as the metric of the joint space $\mathcal{T} \times \mathcal{U}$~\citep{Alvarez20}.

Naturally, the metric on this joint space can be defined as $d_{\mathcal{T}\mathcal{U}}((t,u),(t',u')) = \left( d_{\mathcal{T}}(t,t')^{p} + d_{\mathcal{U}}(u,u')^{p} \right)^{1/p}$, for $p \geq 1$. However, in most applications $d_{\mathcal{T}}$ is readily available, while $d_{\mathcal{U}}$ might be scarce, especially in supervised learning (SL) between labels from unrelated label sets \citep{Alvarez20}. Further, we want $d_|mathcal{T}$ and $d_{\mathcal{U}}$ to have the same units to be addable. To overcome these issues, $d_{\mathcal{U}}$ is expressed in terms of $d_{\mathcal{T}}$ by mapping labels $u$ to distributions over the feature space $\mathcal{P(T)}$ as $u \to \alpha_{u}(T) \triangleq P(T \mid U = u) \in \mathcal{P(T)}.$ 
Therefore, the distance between the labels $u$ and $u'$ is defined as the p-Wasserstein distance between $\alpha_{u}(T)$ and $\alpha_{u'}(T)$,
\begin{equation}\label{label_distr_map_2}
\begin{aligned}
    d_{\mathcal{U}}(u,u') &= \mathcal{W}^{p}_{p}(\alpha_{u}(T),\alpha_{u'}(T)) \\
    &= \min_{\pi \in \Pi(\alpha_{u},\alpha_{u'})} \int_{\mathcal{T} \times \mathcal{T}} (d_{\mathcal{T}}(t,t'))^{p} \,d\pi(t,t')\
\end{aligned}
\end{equation}

The metric on the joint space becomes, 
\begin{equation}\label{joint_space_metric}
\begin{aligned}
    &d_{\mathcal{T}\mathcal{U}}((t,u),(t',u')) = \left( d_{\mathcal{T}}(t,t')^{p} + \mathcal{W}^{p}_{p}(\alpha_{u}(T),\alpha_{u'}(T)) \right)^{1/p} 
\end{aligned}
\end{equation}

Let $\mathcal{Z} = \mathcal{T} \times \mathcal{U}$, then the p-Wasserstein distance between $\hat{P}_{A}(t,u)$ and $\hat{P}_{B}(t,u)$ is a "nested" Wasserstein distance: 
\begin{equation}\label{pre_otdd}
\begin{aligned}
    \mathcal{W}^{p}_{p}(\hat{P}_{A},\hat{P}_{B})
    &= \min_{\pi \in \Pi(P_{A},P_{B})} \int_{\mathcal{Z} \times \mathcal{Z}} (d_{\mathcal{Z}}(z,z'))^{p} \,d\pi \\
    &= \min_{\pi \in \Pi(P_{A},P_{B})} \int_{\mathcal{TU} \times \mathcal{TU}} \left(d_{\mathcal{T}}(t,t')^{p} + \mathcal{W}^{p}_{p}(\alpha_{u},\alpha_{u'}) \right) \,d\pi 
\end{aligned}
\end{equation}

$W^{p}_{p}(\hat{P}_{A},\hat{P}_{B})$ is the OTDD between datasets $\mathcal{D}_{A}$ and $\mathcal{D}_{B}$, often expressed as $d_{OT}(\mathcal{D}_{A},\mathcal{D}_{B})$. This is used in transfer learning to determine the distance (or similarity) between datasets.

\subsection{Proof of Proposition 4}
\label{Appendix prop 4}
We compute the error in occupancy measure for both the infinite and finite horizon cases. In infinite horizon MDPs, the occupancy measure is defined as the expected discounted number of visits of a state-action pair $(s,a)$ in a trajectory \citep{Laroche23}: $\mu = (1 - \gamma) \sum_{t=0}^{\infty} \gamma^t \mu_t$, where $\mu_t = P(s_t,a_t \mid \pi, \eta)$ is the state-action probability distribution at time step $t$ with the initial state distribution $\eta$ following the policy $\pi$. In finite horizon MDPs, the occupancy measure is the expected number of visits of a state-action pair $(s,a)$ in an episode of length $H$ \citep{Altman99}: $\mu = \frac{1}{H} \sum_{t=1}^{H} \mu_t$.



First, we derive error bounds for the infinite horizon MDP in which $\gamma < 1$ and the occupancy measure is approximated using a finite number of samples collected up to a finite number of time steps $T$. Later, we derive error bounds for the finite horizon MDP. 

\subsubsection{Infinite Horizon MDPs}

\textbf{Estimated Occupancy Measure.} 
For convenience, we express the occupancy measure as $\mu = (1 - \gamma) \sum_{t=0}^{\infty} \gamma^t \mu_t$, where $\mu_t = P(s_t,a_t \mid \pi, \eta)$ is the state-action probability distribution at time step $t$ with the initial state distribution $\eta$ following the policy $\pi$. To compute $\mu$, we roll out $N$ episodes (each of multiple time steps) using $\pi$, and take $N$ number of samples at $t$ to approximate $\mu_t$. Thus, the empirical occupancy measure $\hat{\mu}$ is given by $\hat{\mu} = \rho \sum_{t=0}^{T} \gamma^t \hat{\mu}^{N}_t$, where $\rho = \frac{1}{\sum_{t=0} ^{T} \gamma^{t}}$. Note that the total number of samples in the policy dataset $\mathcal{D}_{\pi}$ is $|\mathcal{D}_{\pi}| = N(T+1)$.

\textbf{Occupancy Measure Estimation Error.} Consider two occupancy measures $\mu = (1 - \gamma) \sum_{t=0}^{\infty} \gamma^t \mu_t$ and $\nu = (1 - \gamma) \sum_{t=0}^{\infty} \gamma^t \nu_t$ (with estimates $\hat{\mu} = \rho  \sum_{t=0}^{T} \gamma^t \hat{\mu}^{N_{\mu}}_t$ and $\hat{\nu} = \rho  \sum_{t=0}^{T} \gamma^t \hat{\nu}^{N_{\nu}}_t$). For independent sets $\{\mu_{t}\}_{t \geq 0}$ and $\{\nu_{t}\}_{t \geq 0}$, the Wasserstein distance has the following additive property \citep{Panaretos19}, 
\begin{equation}\label{additive_property}
\begin{aligned}
  \mathcal{W}_{p}( \sum_{t} \mu_{t}, \sum_{t} \nu_{t}) \leq \sum_{t} \mathcal{W}_{p}(\mu_{t},\nu_{t}) 
\end{aligned}
\end{equation}

While for $a \in \mathbb{R}$ \citep{Panaretos19}, 
\begin{equation}\label{additive_property_constant}
\begin{aligned}
    \mathcal{W}_{p}( a\mu, av) = |a| \mathcal{W}_{p}(\mu,v) 
\end{aligned}
\end{equation} 

Therefore, for our scenario where $p = 1$, the Wasserstein distance between $\mu$ and $\nu$ is given by:
\begin{equation}\label{additive_wasserstein}
\begin{aligned}
  \mathcal{W}_{1}(\mu,\nu) &= \mathcal{W}_{1}( (1 - \gamma) \sum_{t=0}^{\infty} \gamma^t \mu_t, (1 - \gamma) \sum_{t=0}^{\infty} \gamma^t \nu_t) \\
  &\leq (1 - \gamma) \sum_{t=0}^{\infty} \gamma^t \mathcal{W}_{1}(\mu_t,\nu_t)
\end{aligned}
\end{equation}

while for $\hat{\mu}$ and $\hat{\nu}$,
\begin{equation}\label{additive_wasserstein_estimate}
\begin{aligned}
  \mathcal{W}_{1}(\hat{\mu},\hat{\nu}) \leq \rho \sum_{t=0}^{T} \gamma^t \mathcal{W}_{1}(\hat{\mu}^{N_{\mu}}_t,\hat{\nu}^{N_{\nu}}_t)
\end{aligned}
\end{equation}

In the RL problems we consider, the state-action space $\mathcal{Z} = \mathcal{S} \times \mathcal{A}$ is commonly defined as the subset of the Euclidean space $\mathcal{Z} \in \mathbb{R}^{B}$, where usually $B \geq 2$. Theorems 1 and 3 in \citep{Sommerfeld19} establish the following error bounds between the true and empirical probability distributions,
\begin{equation}\label{sommerfeld}
\begin{aligned}
    \mathbb{E}[\mathcal{W}_{1}(\hat{\mu}^{N_{\mu}}_{t},\mu_{t})] \leq \mathcal{E}_{2}N_{\mu}^{-\frac{1}{2}} \\
    \mathbb{E}[\mathcal{W}_{1}(\hat{\nu}^{N_{\nu}}_{t},\nu_{t})] \leq \mathcal{E}_{2}N_{\nu}^{-\frac{1}{2}}
\end{aligned}
\end{equation}

where 
\begin{align*}
\mathcal{E}_{2} &\leq 4B^{1/2} diam(\mathcal{Z}) \cdot
\begin{cases}
     2+(1/2)\text{log}_{2}|\mathcal{Z}| & \text{ if $B=2$} \\
     |\mathcal{Z}|^{1/2 - 1/B} \left[ 2+ 1/(2^{B/2 -1} - 1) \right] & \text{ if $B>2$} 
\end{cases}
\end{align*}
Note that $|\mathcal{Z}|$ and $diam(\mathcal{Z})$ denote the cardinality and diameter of $\mathcal{Z}$, respectively. 

Suppose $a = \mathcal{W}_{1}(\hat{\mu},\hat{\nu})$, $b = \mathcal{W}_{1}(\hat{\mu},\mu)$, $c = \mathcal{W}_{1}(\hat{\nu},\mu)$, $d = \mathcal{W}_{1}(\mu,\nu)$, and $e = \mathcal{W}_{1}(\hat{\nu},\nu)$. Then by performing two reverse triangle inequalities,
\begin{equation} \label{2_triangle_inequality}
\begin{aligned}
    &|a - c| \leq b ~~~\text{ and }~~~    |c - d|\leq e \\
    \implies &|a - d| \leq b + e
\end{aligned}
\end{equation}

\Eqref{2_triangle_inequality} implies that,
\begin{equation}\label{bounds_empirical}
\begin{aligned}
    \mathbb{E} [ | \mathcal{W}_{1}(\hat{\mu},\hat{\nu}) - \mathcal{W}_{1}(\mu,\nu) | ] 
    &\leq \mathbb{E} [ \mathcal{W}_{1}(\hat{\mu},\mu) + \mathcal{W}_{1}(\hat{\nu},\nu) ] \\
    &= \mathbb{E} [ \mathcal{W}_{1}(\rho \sum_{t=0}^T \gamma^t\hat{\mu}^{N_{\mu}}_{t},\mu) + \mathcal{W}_{1}(\rho  \sum_{t=0}^T \gamma^t\hat{\nu}^{N_{\nu}}_{t},\nu) ] \\
    &= \mathbb{E} [ \mathcal{W}_{1}(\rho \sum_{t=0}^T \gamma^t\hat{\mu}^{N_{\mu}}_{t},\mu)] + \mathbb{E} [ \mathcal{W}_{1}(\rho  \sum_{t=0}^T \gamma^t\hat{\nu}^{N_{\nu}}_{t},\nu) ] \\
    &+  \mathbb{E} [\mathcal{W}_{1}( (1-\gamma) \sum_{t=0}^\infty \gamma^t\hat{\mu}^{N_{\mu}}_{t},\mu) -  \mathcal{W}_{1}( (1-\gamma) \sum_{t=0}^\infty \gamma^t\hat{\mu}^{N_{\mu}}_{t},\mu)] \\
    & + \mathbb{E} [\mathcal{W}_{1}((1-\gamma)\sum_{t=0}^\infty \gamma^t\hat{\nu}^{N_{\nu}}_{t},\nu) - \mathcal{W}_{1}((1-\gamma) \sum_{t=0}^\infty \gamma^t\hat{\nu}^{N_{\nu}}_{t},\nu)]
\end{aligned}
\end{equation}

By virtue of triangle inequalities, we get
\begin{equation} \label{triangle_inequality_wasersteins}
\begin{aligned}
    &\mathcal{W}_{1}(\rho \sum_{t=0}^T \gamma^t\hat{\mu}^{N_{\mu}}_{t},(1-\gamma) \sum_{t=0}^\infty \gamma^t\hat{\mu}^{N_{\mu}}_{t}) \geq \mathcal{W}_{1}(\rho \sum_{t=0}^T \gamma^t\hat{\mu}^{N_{\mu}}_{t},\mu) - \mathcal{W}_{1}( (1-\gamma) \sum_{t=0}^\infty \gamma^t\hat{\mu}^{N_{\mu}}_{t},\mu) \\
    &\mathcal{W}_{1}(\rho \sum_{t=0}^T \gamma^t\hat{\nu}^{N_{\nu}}_{t},(1-\gamma) \sum_{t=0}^\infty \gamma^t\hat{\nu}^{N_{\nu}}_{t}) \geq \mathcal{W}_{1}(\rho \sum_{t=0}^T \gamma^t\hat{\nu}^{N_{\nu}}_{t},\nu) - \mathcal{W}_{1}( (1-\gamma) \sum_{t=0}^\infty \gamma^t\hat{\nu}^{N_{\nu}}_{t},\nu)
\end{aligned}
\end{equation}

Therefore, the right-hand-side (R.H.S) of \Eqref{bounds_empirical} can be further simplified as
\begin{equation} \label{bounds_empirical_2}
\begin{aligned}
    &\text{R.H.S} \leq \mathbb{E} [\mathcal{W}_{1}(\rho \sum_{t=0}^T \gamma^t\hat{\mu}^{N_{\mu}}_{t},(1-\gamma) \sum_{t=0}^\infty \gamma^t\hat{\mu}^{N_{\mu}}_{t}) ] +  \mathbb{E} [\mathcal{W}_{1}(\rho \sum_{t=0}^T \gamma^t\hat{\nu}^{N_{\nu}}_{t},(1-\gamma) \sum_{t=0}^\infty \gamma^t\hat{\nu}^{N_{\nu}}_{t}) ] \\
    &+ \mathbb{E} [ \mathcal{W}_{1}((1-\gamma) \sum_{t=0}^\infty \gamma^t\hat{\mu}^{N_{\mu}}_{t},\mu) ] + \mathbb{E} [ \mathcal{W}_{1}((1-\gamma)  \sum_{t=0}^\infty \gamma^t\hat{\nu}^{N_{\nu}}_{t},\nu) ]
\end{aligned}
\end{equation}

For simplicity, we denote $\hat{\mu}_{\infty} = (1-\gamma) \sum_{t=0}^\infty \gamma^t\hat{\mu}^{N_{\mu}}_{t}$ (similarly $\hat{\nu}_{\infty}$) and $\hat{\mu}_T = \rho \sum_{t=0}^T \gamma^t\hat{\mu}^{N_{\mu}}_{t}$ (similarly $\hat{\nu}_T$), where $\rho = \frac{1}{\sum_{t=0}^{T} \gamma^t}= \frac{1-\gamma}{1-\gamma^{T+1}}$. Using Theorem 4 in~\citep{Gibbs02}, the 1-Wasserstein metric $\mathcal{W}_{1}$ and the total variation distance $d_{TV}$ satisfy the following,
\begin{equation} \label{wasserstein_total_variation}
\begin{aligned}
    \mathcal{W}_{1}(\hat{\mu}_{\infty},\hat{\mu}_{T})  &\leq  diam(\mathcal{Z}) \cdot d_{TV}(\hat{\mu}_{\infty},\hat{\mu}_{T}) \\
    &= diam(\mathcal{Z}) \cdot \frac{1}{2}\sum_{z \in \mathcal{Z}}|\hat{\mu}_{\infty}(z) - \hat{\mu}_{T}(z)| 
\end{aligned} 
\end{equation}
However, 
\begin{equation} \label{difference_approx}
\begin{aligned}
    \hat{\mu}_{\infty} - \hat{\mu}_{T} &= (1-\gamma) \sum_{t=0}^\infty \gamma^t\hat{\mu}^{N_{\mu}}_{t} - \frac{1-\gamma}{1-\gamma^{T+1}} \sum_{t=0}^T \gamma^t\hat{\mu}^{N_{\mu}}_{t} \\
    &= (1-\gamma) \sum_{t=0}^\infty \gamma^t\hat{\mu}^{N_{\mu}}_{t} - \frac{1-\gamma}{1-\gamma^{T+1}} \sum_{t=0}^T \gamma^t\hat{\mu}^{N_{\mu}}_{t} \\
    &+ (1-\gamma)\sum_{t=0}^T \gamma^t\hat{\mu}^{N_{\mu}}_{t} - (1-\gamma)\sum_{t=0}^T \gamma^t\hat{\mu}^{N_{\mu}}_{t} \\
    &= (1-\gamma) \left( \sum_{t=0}^\infty \gamma^t\hat{\mu}^{N_{\mu}}_{t} - \sum_{t=0}^T \gamma^t\hat{\mu}^{N_{\mu}}_{t} \right) + \left( (1-\gamma) - \frac{1-\gamma}{1-\gamma^{T+1}}\right) \sum_{t=0}^T \gamma^t\hat{\mu}^{N_{\mu}}_{t}\\
    &= (1-\gamma) \sum_{t=T+1}^\infty \gamma^t\hat{\mu}^{N_{\mu}}_{t} - \gamma^{T+1} \frac{1-\gamma}{1-\gamma^{T+1}} \sum_{t=0}^T \gamma^t\hat{\mu}^{N_{\mu}}_{t} \\
    &\leq (1-\gamma) \sum_{t=T+1}^\infty \gamma^t\hat{\mu}^{N_{\mu}}_{t} \\
    &= \gamma^{T+1} \frac{1-\gamma}{\gamma^{T+1}} \sum_{t=T+1}^\infty \gamma^t\hat{\mu}^{N_{\mu}}_{t} \\
    &= \gamma^{T+1} \hat{\mu}_{T+1,\infty}
\end{aligned} 
\end{equation}

where $\frac{1-\gamma}{\gamma^{T+1}}$ normalizes $\sum_{t=T+1}^\infty \gamma^t\hat{\mu}^{N_{\mu}}_{t}$. We utilize \Eqref{difference_approx} in \Eqref{wasserstein_total_variation} as,
\begin{equation} \label{difference_approx_inequality}
\begin{aligned}
    \mathcal{W}_{1}(\hat{\mu}_{\infty},\hat{\mu}_{T})  &\leq diam(\mathcal{Z}) \cdot \frac{1}{2}\sum_{z \in \mathcal{Z}}|\hat{\mu}_{\infty}(z) - \hat{\mu}_{T}(z)|  \\
    &\leq diam(\mathcal{Z}) \cdot  \frac{1}{2} \sum_{z \in \mathcal{Z}}|\gamma^{T+1} \hat{\mu}_{T+1,\infty}(z)| \\
    &=  \frac{\gamma^{T+1}}{2} diam(\mathcal{Z}) 
\end{aligned} 
\end{equation}

\Eqref{difference_approx_inequality} also applies for $\mathcal{W}_{1}(\hat{\nu}_{\infty},\hat{\nu}_{T})$, therefore by substituting these into \Eqref{bounds_empirical_2},
\begin{equation} \label{bounds_empirical_3}
\begin{aligned}
    \text{R.H.S} &\leq \mathbb{E} [ \mathcal{W}_{1}((1-\gamma) \sum_{t=0}^\infty \gamma^t\hat{\mu}^{N_{\mu}}_{t},\mu) ] + \mathbb{E} [ \mathcal{W}_{1}((1-\gamma)  \sum_{t=0}^\infty \gamma^t\hat{\nu}^{N_{\nu}}_{t},\nu) ] + \gamma^{T+1}diam(\mathcal{Z}) \\
    &= \mathbb{E} [ \mathcal{W}_{1}((1-\gamma) \sum_{t=0}^\infty \gamma^t\hat{\mu}^{N_{\mu}}_{t},(1-\gamma) \sum_{t=0}^\infty \gamma^t \mu_t) ] \\
    &+ \mathbb{E} [ \mathcal{W}_{1}((1-\gamma)  \sum_{t=0}^\infty \gamma^t\hat{\nu}^{N_{\nu}}_{t},(1-\gamma) \sum_{t=0}^\infty \gamma^t \nu_t) ] + \gamma^{T+1}diam(\mathcal{Z}) \\
    &\leq (1-\gamma) \sum_{t=0}^\infty \gamma^t \left(\mathbb{E} [\mathcal{W}_{1}(\hat{\mu}^{N_{\mu}}_{t},\mu_t)] + \mathbb{E} [\mathcal{W}_{1}(\hat{\nu}^{N_{\mu}}_{t}, \nu_t)]\right) + \gamma^{T+1}diam(\mathcal{Z})\,.
\end{aligned}
\end{equation}

By substituting \Eqref{sommerfeld} into \Eqref{bounds_empirical_3}
\begin{equation} \label{bounds_empirical_4}
\begin{aligned}
      \text{R.H.S} &\leq  (1-\gamma) \sum_{t=0}^\infty \gamma^t \left( \mathcal{E}_{2}N_{\mu}^{-\frac{1}{2}} + \mathcal{E}_{2}N_{\nu}^{-\frac{1}{2}} \right) + \gamma^{T+1}diam(\mathcal{Z}) \\
      &=\mathcal{E}_{2} \left( N_{\mu}^{-\frac{1}{2}} + N_{\nu}^{-\frac{1}{2}} \right) + \gamma^{T+1}diam(\mathcal{Z})
\end{aligned}
\end{equation}

Therefore, \Eqref{bounds_empirical} becomes:
\begin{equation} \label{bounds_empirical_complete}
\begin{aligned}
\mathbb{E} [ | \mathcal{W}_{1}(\hat{\mu},\hat{\nu}) - \mathcal{W}_{1}(\mu,\nu) | ] \leq \mathcal{E}_{2} \left( N_{\mu}^{-\frac{1}{2}} + N_{\nu}^{-\frac{1}{2}} \right) + \gamma^{T+1}diam(\mathcal{Z})
\end{aligned}
\end{equation}
\textbf{Over the full trajectory in the occupancy measure space.} The true distance between consecutive policies $\pi_{i}$ and $\pi_{i+1}$ after an update is $\mathcal{W}_{1}(v_{\pi_{i}},v_{\pi_{i+1}})$, which is induced by the $i^{th}$ policy update. We estimate this distance using datasets of the policies, i.e. approximated distributions, using $\mathcal{W}_{1}( \hat{v}_{\pi_i},\hat{v}_{\pi_{i+1}} )$.

For $M$ roll out episodes of each $\pi_{i}$, we use \Eqref{bounds_empirical_complete}, with $N_\mu=N_\nu=M$, to derive the following error bounds, 
\begin{equation}
\begin{aligned}
    \mathbb{E} \left[ \left| \mathcal{W}_{1}(v_{\pi_{i}},v_{\pi_{i+1}}) - \mathcal{W}_{1}( \hat{v}_{\pi_i},\hat{v}_{\pi_{i+1}} ) \right| \right] \leq 2 \mathcal{E}_{2} M^{-\frac{1}{2}}+ \gamma^{T+1}diam(\mathcal{Z})
\end{aligned}
\end{equation}
which is consistent with learning from $\mathcal{D}_{\pi_{i}}$ and then $\mathcal{D}_{\pi_{i+1}}$. By summing sequentially through policies encountered during RL training, we compute the total distance over a path of $N$ segments obtained via policy updates:  
\begin{equation}
\begin{aligned}
    \sum_{i=0}^{N-1} \mathbb{E} \left[ \left| \mathcal{W}_{1}(v_{\pi_{i}},v_{\pi_{i+1}}) - \mathcal{W}_{1}( \hat{v}_{\pi_i},\hat{v}_{\pi_{i+1}} ) \right| \right]  \leq 2 N\mathcal{E}_{2} M^{-\frac{1}{2}}+ N\gamma^{T+1}diam(\mathcal{Z})
\end{aligned}
\end{equation}

Since $|\sum_t x_t| \leq \sum_t|x_t|$ then, 
\begin{equation}\label{bounded}
\begin{aligned}
    \mathbb{E} \left[ \left| \sum_{i=0}^{N-1} \mathcal{W}_{1}(v_{\pi_{i}},v_{\pi_{i+1}}) - \sum_{i=0}^{N-1} \mathcal{W}_{1}( \hat{v}_{\pi_i},\hat{v}_{\pi_{i+1}} ) \right| \right]  \leq \frac{2 N\mathcal{E}_{2}} {\sqrt{M}}+ N\gamma^{T+1}diam(\mathcal{Z})
\end{aligned}
\end{equation}

\subsubsection{Finite Horizon MDPs}

\textbf{Occupancy Measure Estimated Error.} Consider two occupancy measures $\mu = \frac{1}{H} \sum_{t=1}^{H} \mu_t$ and $\nu = \frac{1}{H} \sum_{t=1}^{H} \nu_t$ with estimates $\hat{\mu} = \frac{1}{H} \sum_{t=1}^{H} \hat{\mu}^{N_{\mu}}_t$ and $\hat{\nu} = \frac{1}{H} \sum_{t=1}^{H} \hat{\nu}^{N_{\nu}}_t$. From \Eqref{2_triangle_inequality}, we have

\begin{equation} \label{finite_bounds_empirical}
\begin{aligned}
    &\mathbb{E} [ | \mathcal{W}_{1}(\hat{\mu},\hat{\nu}) - \mathcal{W}_{1}(\mu,\nu) | ] \\
    &\leq \mathbb{E} [ \mathcal{W}_{1}(\hat{\mu},\mu) + \mathcal{W}_{1}(\hat{\nu},\nu) ] \\
    &= \mathbb{E} [ \mathcal{W}_{1}(\frac{1}{H} \sum_{t=1}^H \hat{\mu}^{N_{\mu}}_{t}, \frac{1}{H} \sum_{t=1}^H \mu_t) + \mathcal{W}_{1}(\frac{1}{H} \sum_{t=1}^H \hat{\nu}^{N_{\nu}}_{t}, \frac{1}{H} \sum_{t=1}^H \nu_t) ] \\
    &\leq \frac{1}{H} \sum_{t=1}^H \mathbb{E} [ \mathcal{W}_{1}( \hat{\mu}^{N_{\mu}}_{t},\mu_t) ] + \frac{1}{H} \sum_{t=1}^H \mathbb{E} [ \mathcal{W}_{1}(\hat{\nu}^{N_{\nu}}_{t}, \nu_t) ] \\
    &\le \mathcal{E}_{2} \left( N_{\mu}^{-\frac{1}{2}} + N_{\nu}^{-\frac{1}{2}} \right)
\end{aligned}
\end{equation}

Therefore \textbf{for the total path in the occupancy measure space} with $M$ roll out episodes of each $\pi_{i}$, the error bound is 
\begin{equation}\label{finite_bounded}
\begin{aligned}
    \mathbb{E} \left[ \left| \sum_{i=0}^{N-1} \mathcal{W}_{1}(v_{\pi_{i}},v_{\pi_{i+1}}) - \sum_{i=0}^{N-1} \mathcal{W}_{1}( \hat{v}_{\pi_i},\hat{v}_{\pi_{i+1}} ) \right| \right]  \leq \frac{2 N\mathcal{E}_{2}} {\sqrt{M}}\,
\end{aligned}
\end{equation}
by assigning $N_{\mu}= N_{\nu} = M$ in \Eqref{finite_bounds_empirical}, which concludes the proof.
\subsection{Proof of Proposition~\ref{prop:eta_sub}}
\label{Appendix Inequalities}
By definition of $\eta_{sub}$, we get
\begin{align}
    \eta_{sub} &= \frac{\sum_{i=0}^{N-2}\mathcal{W}_{1}(v_{\pi_i},v_{\pi_{i+1}}) + \mathcal{W}_{1}(v_{\pi_{N-1}}, v_{\pi_{N}})}{\mathcal{W}_{1}(v_{\pi_{0}}, v_{\pi_{N}})} \notag\\
    &= \frac{\sum_{i=0}^{N-2}\mathcal{W}_{1}(v_{\pi_i},v_{\pi_{i+1}}) + \mathcal{W}_{1}(v_{\pi_{N-1}}, v_{\pi_{N}})}{\mathcal{W}_{1}(v_{\pi_{0}}, v_{\pi^*})}\times \frac{\mathcal{W}_{1}(v_{\pi_{0}}, v_{\pi^*})}{\mathcal{W}_{1}(v_{\pi_{0}}, v_{\pi_{N}})} \notag\\
    &\geq \frac{\sum_{i=0}^{N-2}\mathcal{W}_{1}(v_{\pi_i},v_{\pi_{i+1}}) + \mathcal{W}_{1}(v_{\pi_{N-1}}, v_{\pi^*})- \mathcal{W}_{1}(v_{\pi_{N}}, v_{\pi^*})}{\mathcal{W}_{1}(v_{\pi_{0}}, v_{\pi^*})} \times \frac{\mathcal{W}_{1}(v_{\pi_{0}}, v_{\pi^*})}{\mathcal{W}_{1}(v_{\pi_{0}}, v_{\pi_{N}})}\notag\\
    &= \left(\eta - \frac{\mathcal{W}_{1}(v_{\pi_{N}}, v_{\pi^*})}{\mathcal{W}_{1}(v_{\pi_{0}}, v_{\pi_{N}})}\right)\frac{\mathcal{W}_{1}(v_{\pi_{0}}, v_{\pi^*})}{\mathcal{W}_{1}(v_{\pi_{0}}, v_{\pi_{N}})}\,. \label{eq:first_triangle}
\end{align}
The inequality above is true due to the triangle inequality $\mathcal{W}_{1}(v_{\pi_{N-1}},v_{\pi_{N}}) + \mathcal{W}_{1}(v_{\pi_{N}},v_{\pi^{*}}) \geq \mathcal{W}_{1}(v_{\pi_{N-1}},v_{\pi^{*}})$. 

By applying triangle inequality, we also get
\begin{align*}
    \mathcal{W}_{1}(v_{\pi_{0}},v_{\pi^{*}}) + \mathcal{W}_{1}(v_{\pi_{N}},v_{\pi^{*}}) \geq \mathcal{W}_{1}(v_{\pi_{0}},v_{\pi_{N}})\,.
\end{align*}
This implies that
\begin{align}\label{eq:lb_ratio}
    \frac{\mathcal{W}_{1}(v_{\pi_{0}},v_{\pi^{*}})}{\mathcal{W}_{1}(v_{\pi_{0}},v_{\pi_{N}})} \geq 1- \frac{\mathcal{W}_{1}(v_{\pi_{N}},v_{\pi^{*}})}{\mathcal{W}_{1}(v_{\pi_{0}},v_{\pi_{N}})}\,.
\end{align}
\Eqref{eq:first_triangle} and \Eqref{eq:lb_ratio} together yield
\begin{align*}
    \eta_{sub} &\geq \left(\eta - \frac{\mathcal{W}_{1}(v_{\pi_{N}}, v_{\pi^*})}{\mathcal{W}_{1}(v_{\pi_{0}}, v_{\pi_{N}})}\right)\left( 1- \frac{\mathcal{W}_{1}(v_{\pi_{N}},v_{\pi^{*}})}{\mathcal{W}_{1}(v_{\pi_{0}},v_{\pi_{N}})}\right)\\
    &= \eta - \frac{\mathcal{W}_{1}(v_{\pi_{N}}, v_{\pi^*})}{\mathcal{W}_{1}(v_{\pi_{0}}, v_{\pi_{N}})} - \eta \frac{\mathcal{W}_{1}(v_{\pi_{N}}, v_{\pi^*})}{\mathcal{W}_{1}(v_{\pi_{0}}, v_{\pi_{N}})} + \left( \frac{\mathcal{W}_{1}(v_{\pi_{N}}, v_{\pi^*})}{\mathcal{W}_{1}(v_{\pi_{0}}, v_{\pi_{N}})} \right)^2\\
    &\geq \eta \left(1 - \frac{\mathcal{W}_{1}(v_{\pi_{N}}, v_{\pi^*})}{\mathcal{W}_{1}(v_{\pi_{0}}, v_{\pi_{N}})}\right) - \frac{\mathcal{W}_{1}(v_{\pi_{N}}, v_{\pi^*})}{\mathcal{W}_{1}(v_{\pi_{0}}, v_{\pi_{N}})}\\
    &\geq \eta \left(1 - \frac{2\mathcal{W}_{1}(v_{\pi_{N}}, v_{\pi^*})}{\mathcal{W}_{1}(v_{\pi_{0}}, v_{\pi_{N}})}\right)\,.
\end{align*}
The second last inequality is due to non-negativity of $\left( \frac{\mathcal{W}_{1}(v_{\pi_{N}}, v_{\pi^*})}{\mathcal{W}_{1}(v_{\pi_{0}}, v_{\pi_{N}})} \right)^2$. The last inequality is due to the fact that $\eta \ge 1$.

Thus, we conclude that
\begin{align*}
    \frac{\eta - \eta_{sub}}{\eta} \leq \frac{2\mathcal{W}_{1}(v_{\pi_{N}}, v_{\pi^*})}{\mathcal{W}_{1}(v_{\pi_{0}}, v_{\pi_{N}})}\,.
\end{align*}

\newpage

\section{Additional Experimental Analysis and Results}
\subsection{Environment Description}\label{Environment Description}

\textbf{2D-Gridworld} environment of size 5x5 with actions: \{up, right, down, left\}. The start and goal states are always located at top-left and bottom-right, respectively. A) \textit{Deterministic, dense rewards setting}: State transitions are deterministic. The reward function is given by $\lVert s_{t} - s_{g}\rVert_{1}$, where $s_{t}$ is the agent state at timestep $t$ and $s_{g}$ is the goal state. B) \textit{Deterministic, sparse rewards setting}: State transitions are deterministic and all states issue a reward of -0.04 except the goal state with reward of 1. C) \textit{Stochastic, dense rewards setting}: Actions have a probability of 0.8 in the instructed direction and 0.1 in each adjacent direction. Reward function is as defined in setting A. 

\begin{figure}[h!]
    \centering
    \includegraphics[width=0.5\linewidth]{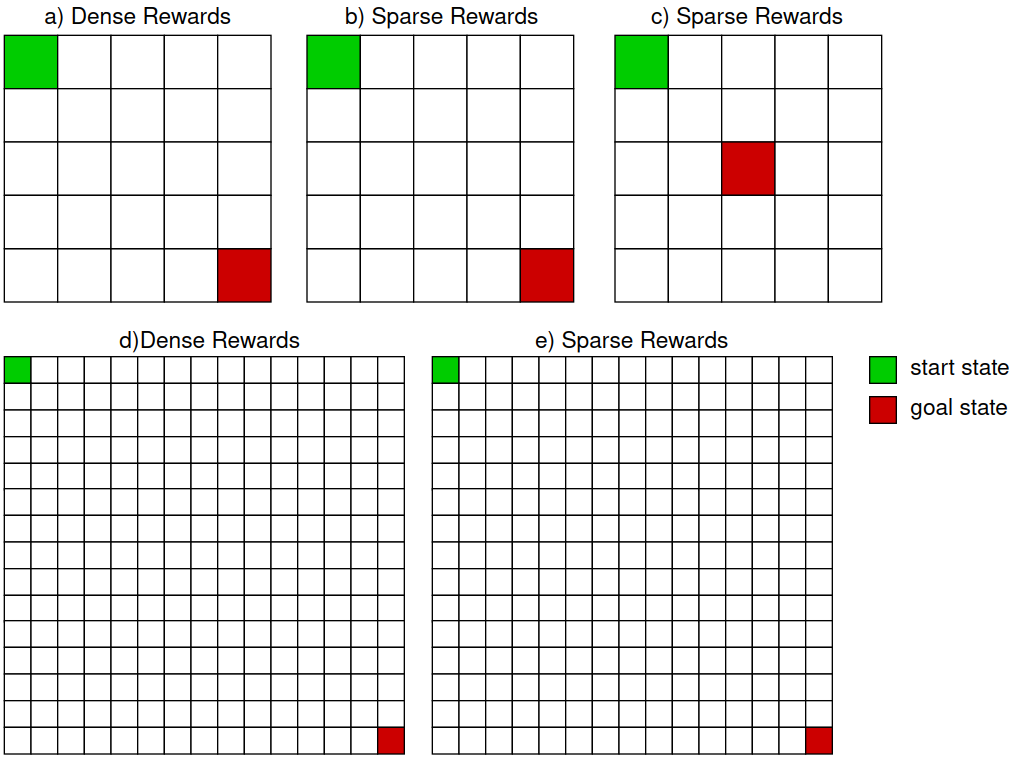}
    \caption{Five gridworld tasks with the same action space, but different rewards, state space and location of the goal state.}\label{fig:task_difficulty}
\end{figure}

\textbf{2D-Gridworlds (Task Difficulty).} Figure~\ref{fig:task_difficulty} depicts the configurations of the 5 tasks that were used to assess ESL with respect to task hardness. They are all deterministic with actions: \{up, right, down, left\}, and mostly have the start-state at the top-left and the goal-state at the bottom-right - with only one task that has the goal-state at the center. In the order of appearance: a) \textit{[5x5] dense}: has size 5x5 and dense rewards, b) \textit{[5x5] sparse (hard)}: has size 5x5 and sparse rewards, c) \textit{[5x5] sparse (easy)}: has size 5x5, sparse rewards, and goal-state at the center, d) \textit{[15x15] dense}: has size 15x15 and dense rewards, and e) \textit{[15x15] sparse}: has size 15x15 and sparse rewards. The reward functions for both dense and sparse rewards are as previously described above for \textbf{2D-Gridworld}. 


\vspace{-0.5em}

\subsection{OMR(k): OMR over number of updates}\label{temporal_OMR}
OMR is defined for the entire policy trajectory by Equation~\ref{optimal_movement_ratio} as, 
\begin{align*}
   \kappa &\triangleq \frac{ \sum_{k \in K^+}\mathcal{W}_{1}(v_{\pi_{k}},v_{\pi_{k+1}}) }{ \sum_{k=0}^{N-1} \mathcal{W}_{1}(v_{\pi_{k}},v_{\pi_{k+1}})}\,.
\end{align*}

To observe how it changes with respect to updates, we compute OMR from update $i$ onwards till the end of the learning trajectory, i.e. over subsequences with a decreasing number of policy updates with increasing $i$, using:   
\begin{equation}
\begin{aligned}
   \kappa (i) &\triangleq \frac{ \sum_{k \in K^+, k \ge i}\mathcal{W}_{1}(v_{\pi_{k}},v_{\pi_{k+1}}) }{ \sum_{k=i}^{N-1} \mathcal{W}_{1}(v_{\pi_{k}},v_{\pi_{k+1}})}, \,\, \text{such that } i \in [0, N-T]
\end{aligned}    
\end{equation}

where $T \approx 0.9N$ to ensure that the last subsequence of policy updates have at least 10\% of the total updates in the trajectory. 

\subsection{Computation of Occupancy Measures}
The finite-horizon occupancy measure is defined as~\citep{Altman99}, 
\begin{align*}
    v^H_{\pi}(s,a) = \frac{1}{H}\sum_{t=1}^{H} \prob(s_{t} = s, a_{t} = a \mid \pi, \mu)
\end{align*}
for which the probability of the state-action pair selected is time-dependent. If we restrict our analysis to stationary policies where $\pi(a_t|s_t) = \pi(a|s)$, then the probability of the state-action pair becomes time-independent and thus
\begin{align*}
    v^H_{\pi}(s,a) = \prob(s,a \mid \pi, \mu)
\end{align*}

This implies that the use of stationary policies in finite-horizon MDPs, as observed in practice with many episodic MDPs~\citep{Memmel22,Aleksandrowicz23,Liu23}, induces stationary occupancy measures - where the expected number of state-action pair visits are independent of the time-step. Work by \citep{Bojun20} provides extensive details about the existence of stationarity in episodic MDPs and shows (in Theorem 4) that
\begin{equation}\label{stationarity}
\begin{aligned}
    \mathbb{E}_{(s,a) \sim v_{\pi}^{H}} \left[ \mathcal{\bar{R}}(s,a) \right] = \frac{\mathbb{E}_{\zeta \sim M_{\pi}} \left[ \sum_{t=1}^{H(\zeta)} R(s_t,a_t) \right]}{\mathbb{E}_{\zeta \sim M_{\pi}} \left[ H(\zeta) \right]}
\end{aligned}
\end{equation}
where $\zeta$ is the episodic state-action pair trajectory, $H(\zeta)$ is the episode length corresponding to $\zeta$, and $M_{\pi}$ is the markov chain induced by policy $\pi$. We verified the correctness of our $v_{\pi}^{H}$ computation by calculating the relative error derived from Equation~\ref{stationarity} to check its validity. The relative error is given as
\begin{equation}\label{stationarity_v2}
\begin{aligned}
    \text{Rel. Error \%\ } = 100*\frac{ \mathbb{E}_{(s,a) \sim v_{\pi}^{H}} \left[ \mathcal{\bar{R}}(s,a) \right] \mathbb{E}_{\zeta \sim M_{\pi}} \left[ H(\zeta) \right] - \mathbb{E}_{\zeta \sim M_{\pi}} \left[ \sum_{t=1}^{H(\zeta)} R(s_t,a_t) \right] }{ \mathbb{E}_{\zeta \sim M_{\pi}} \left[ \sum_{t=1}^{H(\zeta)} R(s_t,a_t) \right] }
\end{aligned}
\end{equation}

\begin{figure}[h!]
\centering\vspace{-0.2em}
\setlength{\arrayrulewidth}{0.1mm}
\setlength{\tabcolsep}{0.0mm}
\begin{tabular}{ll}
\includegraphics[width=0.5\linewidth]{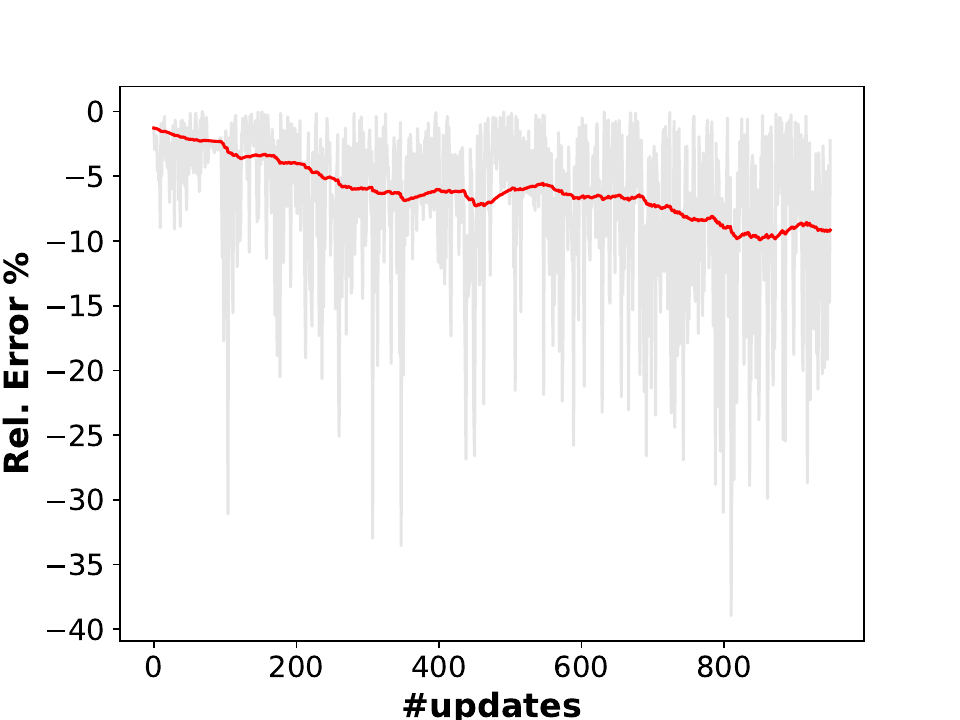} &  \includegraphics[width=0.5\linewidth]{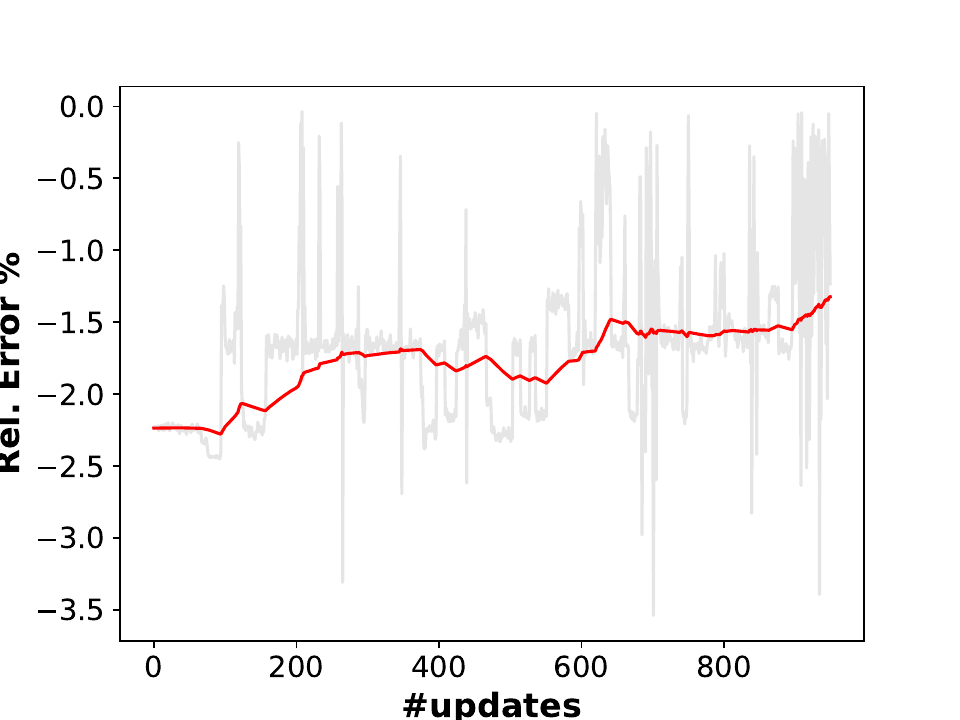}     
\end{tabular}
\caption{Rel. Error$\% $ vs number of updates plots in the 2D-Gridworld environment where $v^H_{\pi}$ is estimated using $M = 10$ rollouts (left) and $M = 500$ rollouts (right). The noisy true data is in grey, while the rolling mean is the red line. }\label{fig:relative_error}\vspace*{-0.5em}
\end{figure}

Figure~\ref{fig:relative_error} depicts Rel. Error$\% $ vs number of updates in the stochastic 2D-Gridworld environment with dense rewards. We observe that increasing the number of rollouts $M$ reduces the estimation error of $v^H_{\pi}$. For $M = 10$, the absolute relative error can be as high as around 40\% with the mean (red line) is at most about 10\%. While for $M = 500$, the maximum absolute relative error is about 3.5\%.

\subsection{Effects of the number of rollouts - SAC}
The policy dataset $D_{\pi_{i}}$ in a deterministic environment is made up of (s,a) pairs generated from a single episode of the policy $\pi_{i}$. In a deterministic environment, this sequence remains the same across repeats of episodes, for each policy $\pi_i$ (deterministic) at update step $i$. Therefore, a single rollout is sufficient to estimate the occupancy measure $v_{\pi_{i}}$. In a stochastic environment, rollouts are impacted by the environment's stochasticity. Thus, multiple rollouts are needed to estimate the occupancy measure accurately. As the number of rollouts increases, the occupancy measure should converge and become less noisy. 

Table~\ref{tab:sac_mult} shows that, in a stochastic setting, the ESL values converge as the number of rollouts increases. OMR appears to be invariant across various the number of rollouts, and the mean number of updates appear to be consistent around 2900 (with exception for \#rollouts = 1). The results indicate that from about 6 rollouts, the estimated occupancy measures become less noisy. This aligns with \Eqref{bounded}, which shows that increasing the number of rollouts reduces estimation error.    

\setlength{\arrayrulewidth}{0.5mm}
\setlength{\tabcolsep}{2.5mm}
\begin{table}[ht!]
\centering\vspace{-0.2em}
\begin{tabular}{lcccr}
\hline
\#rollouts & ESL & OMR & UC \\
\hline
1   & 849.1$\pm$468.5 & 0.500$\pm$0.004 & 2349$\pm$742.2 \\
3 & 618.6$\pm$257.3 & 0.501$\pm$0.005 & 2913$\pm$1397 \\
6  & 445.4$\pm$245.8 & 0.501$\pm$0.042 & 2962$\pm$2043  \\
9   & 428.1$\pm$234.4 & 0.503$\pm$0.004 & 2781$\pm$1743  \\
\hline
\end{tabular}
\caption{Evaluation of SAC algorithm in the \textbf{stochastic, dense-rewards setting} for 5x5 gridworld with \textbf{40 maximum steps per episode} across various number of rollouts. The effects of \#rollouts on the Effort of Sequential Learning (ESL), Optimal Movement Ratio (OMR), and number of updates to convergence (UC) are observed.}
\label{tab:sac_mult}\vspace{-0.5em}
\end{table}

\subsection{$\eta$ vs $\eta_{sub}$}\label{Appendix B.2}
We compare ESL when the optimal policy was reached, denoted $\eta$, versus when it was not, denoted $\eta_{sub}$, in Tables \ref{tab:esl_vs1} and \ref{tab:esl_vs2}. First, we observe that the number of rollouts impacts the metric values. Second, $\eta_{sub}$ values are always greater than $\eta$ values. Note that UCRL2 and PSRL update their policies only at the end of each episode, whereas SAC and DQN update theirs after each time step. Hence, UC$_{sub} = 499$ for both UCRL2 and PSRL.   

The ESL values (both $\eta$ and $\eta_{sub}$) in Table \ref{tab:esl_vs2} are lower than those in Table \ref{tab:esl_vs1}, as expected since more data samples reduce estimation error. The distance from the initial policies to the final polices are not so different. Using both Tables \ref{tab:esl_vs1} and \ref{tab:esl_vs2}, we notice that comparing algorithms with $\eta_{sub}$ yields the same efficiency ranking (e.g. PSRL, UCRL, SAC and DQN) as $\eta$. This indicates that $\eta_{sub}$ reliably predicts results provided by $\eta$ for comparing algorithms.   

The results presented in Table~\ref{tab:env23} for stochastic dense-rewards setting are consistent with those in Table~\ref{tab:esl_vs2} because the number of rollouts used was Nr = 6. 

\setlength{\tabcolsep}{0.8mm}
\begin{table}[ht!]
\centering
\begin{tabular}{l|c|c|c|c|c|r}
\hline
Algo. & $\eta$ & $\eta_{sub}$ & d & c & UC & UC$_{sub}$\\
\hline
SAC   & \shortstack{849$\pm$ \\468} & \shortstack{3623$\pm$ \\4166} & \shortstack{5.63$\pm$ \\1.50}  & \shortstack{5.26$\pm$ \\2.10} & \shortstack{2350$\pm$ \\742} & \shortstack{7951$\pm$ \\3535} \\
\hline
UCRL2 & \shortstack{230$\pm$ \\155} & \shortstack{613$\pm$ \\999} & \shortstack{5.65$\pm$ \\0.93}  & \shortstack{5.45$\pm$ \\2.15} & \shortstack{284$\pm$ \\180} & \shortstack{499$\pm$ \\0.0} \\
\hline
PSRL  & \shortstack{86.2$\pm$ \\44.4} & \shortstack{389$\pm$ \\102} & \shortstack{4.96$\pm$ \\1.26}  & \shortstack{5.29$\pm$ \\1.49} & \shortstack{97.2$\pm$ \\52.5} & \shortstack{499$\pm$ \\0.0} \\
\hline
DQN   & \shortstack{564$\pm$ \\478} & \shortstack{3911$\pm$ \\1710} & \shortstack{5.52$\pm$ \\1.39}  & \shortstack{6.54$\pm$ \\2.05} & \shortstack{1278$\pm$ \\1061} & \shortstack{9162$\pm$ \\1904} \\
\hline
\end{tabular}
\caption{Evaluation of algorithms in the \textbf{stochastic, dense-rewards setting} for 5x5 gridworld with \textbf{40 maximum steps per episode} with the number of rollouts Nr = 1. The total number of training episodes is 500. When the algorithm converged at optimality, $\eta$ is the \textit{Effort of Sequential Learning}, $d = \mathcal{W}_{1}(\pi_0, \pi^*)$ is distance from initial policy to the optimal policy, and UC is the number of updates to convergence. When the algorithm did not converge at the optimal policy, rather a non-optimal $\pi_{N}$, we use $\eta_{sub}$, $c = \mathcal{W}_{1}(\pi_0, \pi_N)$, and UC$_{sub}$ to denote the aforementioned quantities. 40 training trials were used.}
\label{tab:esl_vs1}
\end{table}

\setlength{\tabcolsep}{0.8mm}
\begin{table}[ht!]
\centering
\begin{tabular}{l|c|c|c|c|c|r}
\hline
Algo. & $\eta$ & $\eta_{sub}$ & d & c & UC & UC$_{sub}$\\
\hline
SAC   & \shortstack{445$\pm$ \\246} & \shortstack{853$\pm$ \\127} & \shortstack{5.63$\pm$ \\1.23}  & \shortstack{7.26$\pm$ \\1.45} & \shortstack{2963$\pm$ \\2043} & \shortstack{6793$\pm$ \\441} \\
\hline
UCRL2 & \shortstack{198$\pm$ \\121} & \shortstack{510$\pm$ \\274} & \shortstack{5.36$\pm$ \\0.84}  & \shortstack{4.58$\pm$ \\1.90} & \shortstack{268$\pm$ \\155} & \shortstack{499$\pm$ \\0.0} \\
\hline
PSRL  & \shortstack{55.4$\pm$ \\33.6} & \shortstack{361$\pm$ \\43.6} & \shortstack{4.97$\pm$ \\1.34}  & \shortstack{3.91$\pm$ \\0.48} & \shortstack{76.1$\pm$ \\50.6} & \shortstack{499$\pm$ \\0.0} \\
\hline
DQN   & \shortstack{458$\pm$ \\311} & \shortstack{1971$\pm$ \\250} & \shortstack{4.88$\pm$ \\1.06}  & \shortstack{6.52$\pm$ \\0.31} & \shortstack{1651$\pm$ \\1077} & \shortstack{13778$\pm$ \\6907} \\
\hline
\end{tabular}
\caption{Evaluation of algorithms in the \textbf{stochastic, dense-rewards setting} for 5x5 gridworld with \textbf{40 maximum steps per episode} with the number of rollouts Nr = 6. The total number of training episode is 500. When the algorithm converged at optimality, $\eta$ is the \textit{Effort of Sequential Learning}, $d = \mathcal{W}_{1}(\pi_0, \pi^*)$ is distance from initial policy to the optimal policy, and UC is the number of updates to convergence. When the algorithm did not converge at the optimal policy however some $\pi_{N}$, we use $\eta_{sub}$, $c = \mathcal{W}_{1}(\pi_0, \pi_N)$, and UC$_{sub}$ to denote the aforementioned quantities. 40 training trials were used.}
\label{tab:esl_vs2}
\end{table}

\subsection{Results for over 500 rollouts}\label{Appendix B.3}
In a stochastic environment, the optimal policy $\pi^*$ may not always return optimal rewards due to the stochasticity of the environment. However, we expect the optimal policy to have higher success rate of reaching optimal rewards than any other policy. Therefore, we reason that if the final policy $\pi$ has less than 30\% success rate, then it is considered a sub-optimal policy. In this section, we present more results to compare algorithms in the gridworld with stochastic transitions and dense rewards. The results (shown in Tables~\ref{tab:env_roll500_1} and~\ref{tab:env_roll500_2}) are for over 500 number of rollouts across 10 trial runs.   

\begin{table}[H]
\centering
\begin{tabular}{lcccr}
\hline
Algo.& ESL & OMR & UC & SR\%\\
\hline
SAC   & 97.6$\pm$34 & 0.506$\pm$0.01 & 2130$\pm$328 & 80 \\
UCRL2 & {293}$\pm${89.5}$^{\star}$ & 0.502$\pm$0.02 & 388$\pm$118 & 60\\
PSRL  & \textbf{73.4}$\pm$\textbf{34.2} & 0.49$\pm$0.04 & \textbf{85.5}$\pm$\textbf{40.4} & 80 \\
DQN   & 104$\pm$33.4 & \textbf{0.511}$\pm$\textbf{0.012} & 1069$\pm$334 & 90 \\
\hline
\end{tabular}
\caption{Evaluation of RL algorithms (10 runs, number of rollouts 500) in the \textbf{stochastic, dense-rewards setting} for 5x5 gridworld, including Effort of Sequential Learning (ESL), Optimal Movement Ratio (OMR), number of updates to convergence (UC), and success rate (SR). Lowest ESL, highest OMR and lowest UC values are in \textbf{bold}, while the highest ESL value is starred ($\star$). 
}\label{tab:env_roll500_1}
\end{table}

\setlength{\tabcolsep}{0.8mm}
\begin{table}[ht!]
\centering
\begin{tabular}{l|c|c|c|c|c|r}
\hline
Algo. & $\eta$ & $\eta_{sub}$ & d & c & UC & UC$_{sub}$\\
\hline
SAC   & \shortstack{97.6$\pm$ \\34} & \shortstack{356$\pm$ \\24.3} & \shortstack{4.82$\pm$ \\1.04}  & \shortstack{4.05$\pm$ \\3.04} & \shortstack{2130$\pm$ \\328} & \shortstack{8890$\pm$ \\8192} \\
\hline
UCRL2 & \shortstack{293$\pm$ \\89.5} & \shortstack{449$\pm$ \\208} & \shortstack{5.08$\pm$ \\0.10}  & \shortstack{4.66$\pm$ \\1.59} & \shortstack{388$\pm$ \\118} & \shortstack{499$\pm$ \\0.0} \\
\hline
PSRL  & \shortstack{73.4$\pm$ \\34.2} & \shortstack{67.0$\pm$ \\49.2} & \shortstack{4.32$\pm$ \\1.22}  & \shortstack{5.33$\pm$ \\0.07} & \shortstack{85.5$\pm$ \\40.4} & \shortstack{109$\pm$ \\90.5} \\
\hline
DQN   & \shortstack{104$\pm$ \\33.4} & \shortstack{70.4$\pm$ \\0} & \shortstack{5.54$\pm$ \\0.68}  & \shortstack{6.67$\pm$ \\0} & \shortstack{1069$\pm$ \\335} & \shortstack{950$\pm$ \\0} \\
\hline
\end{tabular}
\caption{Evaluation of algorithms in the \textbf{stochastic, dense-rewards setting} for 5x5 gridworld with \textbf{40 maximum steps per episode} with the number of rollouts is 500. The total number of training episode is 500. When the algorithm converged at optimality, $\eta$ is the \textit{Effort of Sequential Learning}, $d = \mathcal{W}_{1}(\pi_0, \pi^*)$ is distance from initial policy to the optimal policy, and UC is the number of updates to convergence. When the algorithm did not converge at the optimal policy however some $\pi_{N}$, we use $\eta_{sub}$, $c = \mathcal{W}_{1}(\pi_0, \pi_N)$, and UC$_{sub}$ to denote the aforementioned quantities. 40 training trials were used.}
\label{tab:env_roll500_2}
\end{table}

\subsection{Effects of Hyperparameters - UCRL2}
Table~\ref{tab:ucrl_hyp} illustrates the effects of hyperparameter values in the UCRL2 algorithm. The environment is deterministic dense-rewards setting with 200 training episodes. We observe that high exploration rates ($\delta \to 0$) appear to align with high ESL and UC, while high exploitation rates ($\delta \to 1$) appear to align with low ESL and UC. OMR appears to be invariant across various $\delta$ values. 

\setlength{\arrayrulewidth}{0.5mm}
\setlength{\tabcolsep}{1.5mm}
\begin{table}[h!]
\centering
\begin{tabular}{lcccr}
\hline
$\delta$ & ESL & OMR & UC & SR\%\\
\hline
0.1   & 47.76$\pm$7.768 & 0.512$\pm$0.033 & 62.26$\pm$9.977 & 100 \\
0.3 & 39.29$\pm$5.860 & 0.515$\pm$0.034 & 58.08$\pm$7.746 & 100\\
0.5  & 38.26$\pm$6.747 & 0.511$\pm$0.036 & 56.92$\pm$9.111 & 100 \\
0.7   & 37.48$\pm$5.094 & 0.507$\pm$0.029 & 56.68$\pm$7.460 & 100 \\
0.9   & 36.40$\pm$5.301 & 0.510$\pm$0.036 & 54.86$\pm$7.326 & 100 \\
\hline
\end{tabular}
\caption{Evaluation of UCRL2 algorithm in the \textbf{deterministic, dense-rewards setting} for 5x5 gridworld with \textbf{15 maximum steps per episode}. Different confidence parameter $\delta \in (0,1)$ were evaluated to see their effects on Effort of Sequential Learning (ESL), Optimal Movement Ratio (OMR), number of updates to convergence (UC), and success rate (SR). Note that as $\delta \to 0$, the agent approaches absolute exploration, and with $\delta \to 1$ absolute exploitation.}
\label{tab:ucrl_hyp}
\end{table}

\newpage
\section{Specifications of the RL Algorithms under Study}
\subsection{Methods for simulation results (Discrete MDP)}

\textbf{Model parameter initialisation.}
We initialised model parameters for deep learning RL algorithms like DQN and SAC by uniformly sampling weight values between $-3\cdot10^{-4}$ and $3\cdot10^{-4}$ and the biases at $0$. For tabular Q-learning algorithms, we randomly initialized the Q-values between $-1.0$ and $1.0$. For UCRL and PSRL, the policy model was randomly initialized. Note that all Wasserstein distances were computed using a python package POT~\citep{Flamary21}.    

\textbf{Results in Figure \ref{policy_evolution}.} The problem setting was deterministic with dense-rewards and 15 maximum number of steps per episode. The total number of episodes was 200. The convergence criterion was satisfied when maximum returns were produced by an algorithm over 5 consecutive updates. The results showcase a single representative run of each algorithm. The confidence parameter $\delta = 0.1$ was utilized for UCRL2. The $\alpha$ parameter for SAC was autotuned using the approach in \citep{Haarnoja19} along with hyperparameters described in Table \ref{sac}. While DQN began with $\epsilon = 1.0$ and the value decayed as $\epsilon[t+1] = \max\{ 0.9999 \times \epsilon[t], 0.0001  \}$. Table \ref{dqn} shows hyperparameters for DQN. Note that the ADAM~\citep{Kingma17} optimizer was used in all the neural network models. 

\begin{table}[ht]
\caption{SAC Hyperparameters.}\label{sac}
\centering\vspace{-0.2em}
\begin{tabular}{l|cccr}
\hline
Parameter & Value \\
\hline
learning rate & $5\cdot10^{-4}$ \\
discount($\gamma$) & 0.99 \\
replay buffer size & $10^{4}$   \\
number of hidden layers (all networks)  & 1 \\
number of hidden units per layer & 32 \\
number of samples per minibatch & 64 \\
nonlinearity & ReLU \\
entropy target  & -4 \\
target smoothing coefficient ($\tau$) & 0.01 \\
target update interval & 1 \\
gradient steps & 1 \\
initial exploration steps \\ before model starts updating & 500 \\
\hline
\end{tabular}\vspace{-0.5em}
\end{table}

\begin{table}[ht]
\caption{DQN Hyperparameters.}\label{dqn}
\centering\vspace{-0.2em}
\begin{tabular}{l|cccr}
\hline
Parameter & Value \\
\hline
learning rate & $5\cdot10^{-2}$ \\
discount($\gamma$) & 0.99 \\
replay buffer size & $10^{4}$   \\
number of hidden layers (all networks)  & 1 \\
number of hidden units per layer & 32 \\
number of samples per minibatch & 64 \\
nonlinearity & ReLU \\
target smoothing coefficient ($\tau$) & 0.001 \\
target update interval & 1 \\
gradient steps & 1 \\
initial exploration steps \\ before $\epsilon$ decays & 500 \\
\hline
\end{tabular}\vspace{0.0em}
\end{table}

\textbf{Results in Tables \ref{tab:env1} and \ref{tab:env23}.} The problem settings had 40 maximum number of steps per episode, and the convergence criterion was satisfied when maximum returns were produced by an algorithm over 5 consecutive updates. The means and standard deviations for each algorithm were computed over 50 runs. The total number of episodes was 200 for results in Table \ref{tab:env1}, and 500 in Table \ref{tab:env23}. For results in Figure~\ref{envir}, the Q-learning with decaying $\epsilon$-greedy where $\epsilon = 0.9$ was employed in the gridworld tasks described in Appendix~\ref{Environment Description}. A convergence criterion of 50 consecutive model updates with maximum returns was utilized. We aggregated the result over 40 training trials and the maximum number of steps per episode was 60.

\subsection{Methods for simulation results (Continuous MDP)}
\textbf{Model parameter initialisation.} We initialised model parameters for the deep learning SAC algorithm by uniformly sampling weight values between $-3\cdot10^{-4}$ and $3\cdot10^{-4}$ and the biases at $0$. For the DDPG algorithm, the output layer weight values were initialised using Xavier Initialization~\citep{Glorot10}, while the rest were uniformly sampled between $-3\cdot10^{-3}$ and $3\cdot10^{-3}$. This was done on both the actor and critic networks. The ADAM~\citep{Kingma17} optimizer was used in all the neural network models. In both algorithms, 1) a discount factor $\gamma = 0.99$ was used, 2) 500 initial steps were taken before updating model weights, and 3) replay buffer size was $10^6$. Tables~\ref{ddpg} and~\ref{sac_con} display hyperparameters for DDPG and SAC, respectively.  

\textbf{Results in Figure \ref{contin_policy_evolution}.} The problem setting was Mountain Car continuous~\citep{Moore90} with 999 maximum number of steps per episode~\citep{Brockman16}. The total number of training episodes was 100. The convergence criterion was satisfied when maximum returns were produced by an algorithm over 10 consecutive updates. The results showcase a single representative run of each algorithm. For results in \textbf{Table \ref{tab:continuous}}, the mean and standard deviations for each algorithm were computed over 5 runs. While RL training was conducted in a continuous state-action space, we discretized it for Wasserstein distance calculations between occupancy measures, using 4 bins for actions and 10 bins for states. 

\begin{table}[h!]
\caption{DDPG Hyperparameters.}\label{ddpg}
\centering\vspace{-0.2em}
\begin{tabular}{l|cccr}
\hline
Parameter & Value \\
\hline
number of samples per minibatch & 128 \\
nonlinearity & ReLU \\
target smoothing coefficients ($\tau$) & 0.001 \\
target update interval & 1 \\
gradient steps & 1 \\
number of hidden layers (all networks)  & 2 \\
number of hidden units per layer & 64 \\
Actor learning rate & $5\cdot10^{-4}$ \\
Critic learning rate & $5\cdot10^{-3}$ \\
\hline
\end{tabular}\vspace{-0.5em}
\end{table}

\begin{table}[h!]
\caption{SAC Hyperparameters.}\label{sac_con}
\centering\vspace{-0.2em}
\begin{tabular}{l|cccr}
\hline
Parameter & Value \\
\hline
learning rate & $3\cdot10^{-3}$ \\
number of hidden layers (all networks)  & 2 \\
number of hidden units per layer & 64 \\
number of samples per minibatch & 128 \\
nonlinearity & ReLU \\
target smoothing coefficient ($\tau$) & 0.001 \\
target update interval & 1 \\
gradient steps & 1 \\
\hline
\end{tabular}\vspace{0.0em}
\end{table}

\newpage

\section{Supplementary Results}\label{Supplementary Results}
In this section we present enlarged versions of results in Figure~\ref{policy_evolution} (see Section~\ref{More Results}) and additional plots that support the results in the main paper (see Section~\ref{Additional Results}). Note that the Github repository of the project is available at \url{https://github.com/nkhumise-rea/analysis_of_occupancy_measure_trajectory}. 

\subsection{Enlarged Visualisation of the Occupancy Measure Trajectories}\label{More Results}
\begin{figure}[h!]
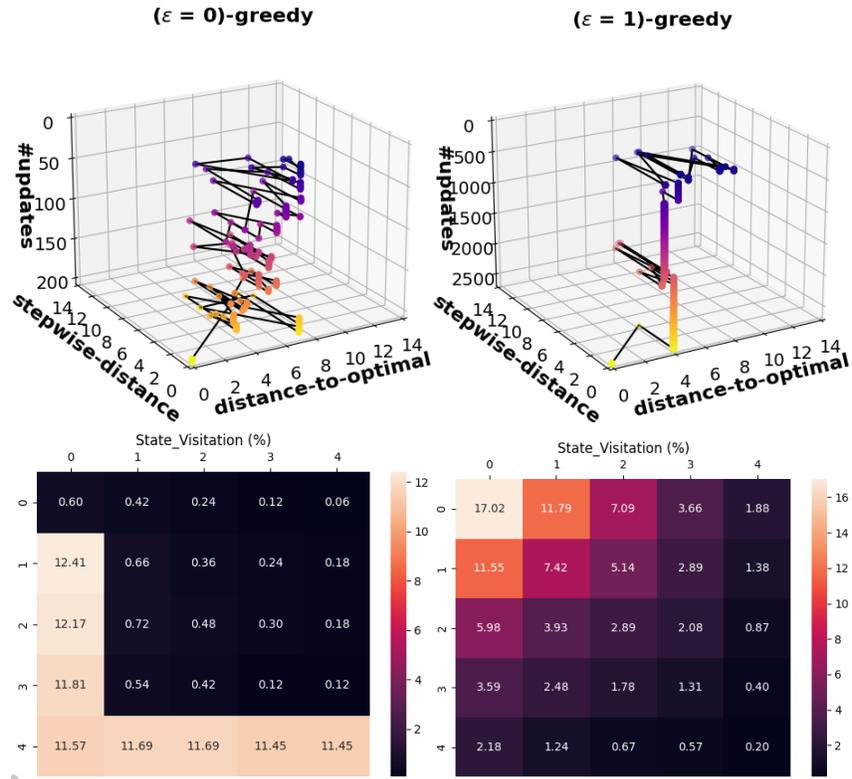

\centering\vspace*{0em}
\setlength{\arrayrulewidth}{0.1mm}
\setlength{\tabcolsep}{0.0mm}
\begin{tabular}{ll}
\includegraphics[width=0.4\linewidth]{images/policy_evolution/greedy.png} &
\includegraphics[width=0.4\linewidth]{images/policy_evolution/random1.png} \\ 
\includegraphics[width=0.4\linewidth]{images/state_visits/greedy_stvt.png} &      \includegraphics[width=0.4\linewidth]{images/state_visits/random_stvt.png}  
\end{tabular}
\caption{Top row: Scatter plots of \textit{distance-to-optimal} and \textit{stepwise-distance} over updates for $\epsilon($=0)-greedy and $\epsilon($=1)-greedy Q-learning. Bottom row: State visitations.}
\label{greedy_random_policy_evolution}\vspace*{-0.5em}
\end{figure}

\begin{figure}[ht!]
\centering\vspace*{-0.2em}
\setlength{\arrayrulewidth}{0.1mm}
\setlength{\tabcolsep}{0.0mm}
\begin{tabular}{ll}
\includegraphics[width=0.4\linewidth]{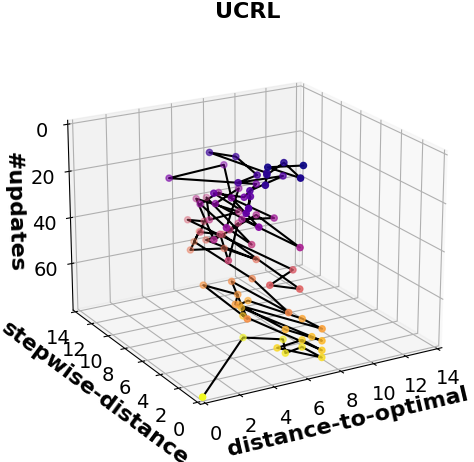} &  \includegraphics[width=0.4\linewidth]{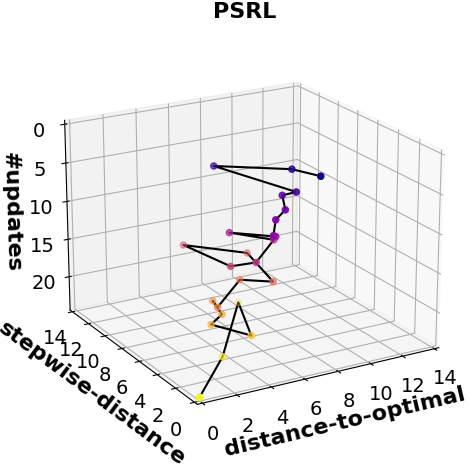}  \\
\includegraphics[width=0.4\linewidth]{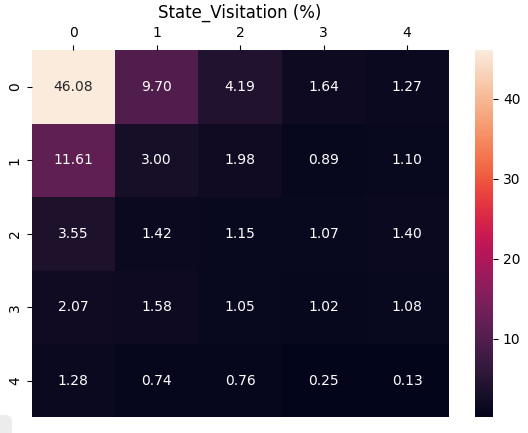} &  \includegraphics[width=0.4\linewidth]{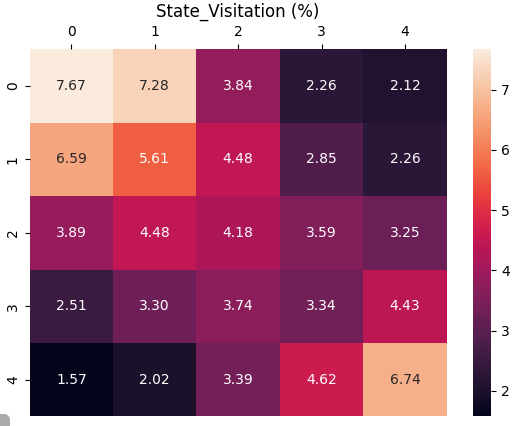}  
\end{tabular}
\caption{Top row: Scatter plots of \textit{distance-to-optimal} and \textit{stepwise-distance} over updates for UCRL and PSRL. Bottom row: State visitations.}
\label{ucrl_psrl_policy_evolution}\vspace*{-0.5em}
\end{figure}

\begin{figure}[ht!]
\centering\vspace*{-0.2em}
\setlength{\arrayrulewidth}{0.1mm}
\setlength{\tabcolsep}{0.0mm}
\begin{tabular}{ll}   
\includegraphics[width=0.45\linewidth]{images/policy_evolution/sac.png} &   \includegraphics[width=0.45\linewidth]{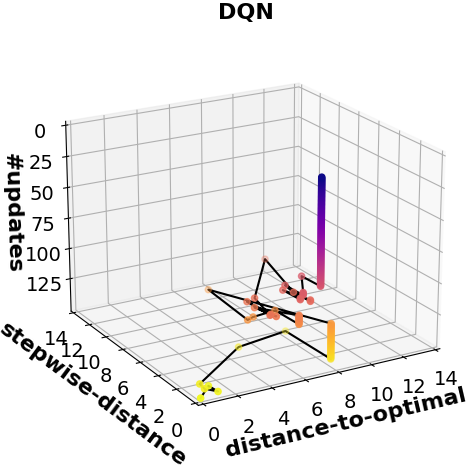} \\
\includegraphics[width=0.45\linewidth]{images/state_visits/sac_stvt.png} &  \includegraphics[width=0.45\linewidth]{images/state_visits/dqn_stvt.png} 
\end{tabular}
\caption{Top row: Scatter plots of \textit{distance-to-optimal} and \textit{stepwise-distance} over updates for SAC and DQN. Bottom row: State visitations.}
\label{sac_dqn_policy_evolution}
\end{figure}

Figures~\ref{greedy_random_policy_evolution} - \ref{sac_dqn_policy_evolution} are enlarged versions of enlarged versions of Figure~\ref{policy_evolution}. For each algorithm, there is a visualisation of the policy trajectory and visualisation of the state visitation below it.    

\clearpage

\subsection{Evolution of  \textit{distance-to-optimal}, \textit{stepwise-distance}, and OMR$(k)$}\label{Additional Results}
\begin{figure}[h!]
\centering\vspace*{0em}
\setlength{\arrayrulewidth}{0.1mm}
\setlength{\tabcolsep}{0.0mm}
\begin{tabular}{ll}
\includegraphics[width=0.45\linewidth]{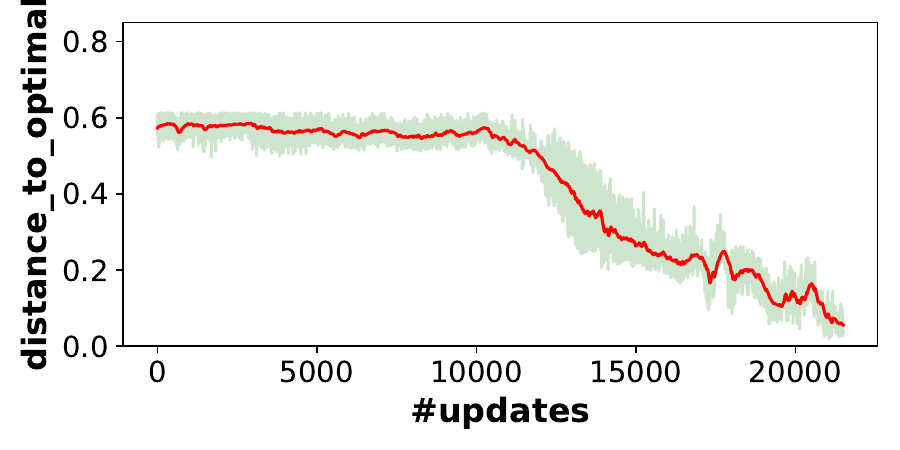} &  \includegraphics[width=0.45\linewidth]{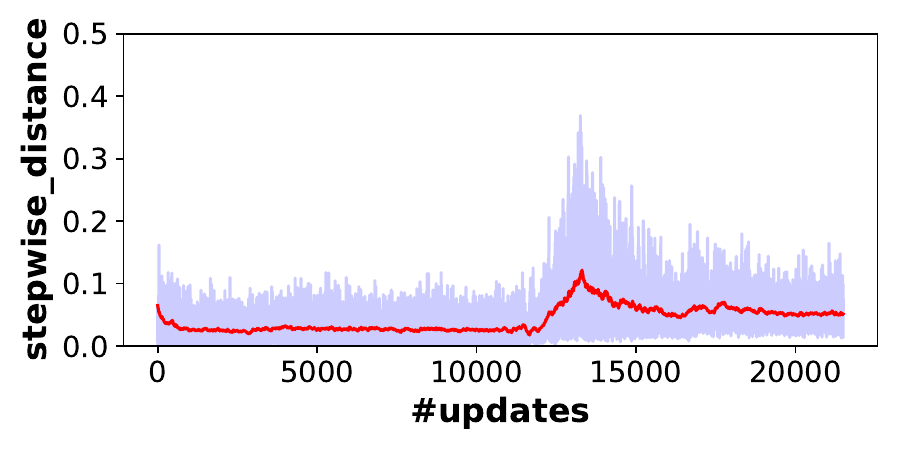}      
\end{tabular}
\caption{\textit{distance-to-optimal} and \textit{stepwise-distance} vs updates plots for SAC algorithm in the Mountain Car continuous environment.}\label{sac_policy_evolution}\vspace*{0.5em}
\end{figure}

\begin{figure}[h!]
\centering\vspace*{0em}
\setlength{\arrayrulewidth}{0.1mm}
\setlength{\tabcolsep}{0.0mm}
\begin{tabular}{ll}
\includegraphics[width=0.45\linewidth]{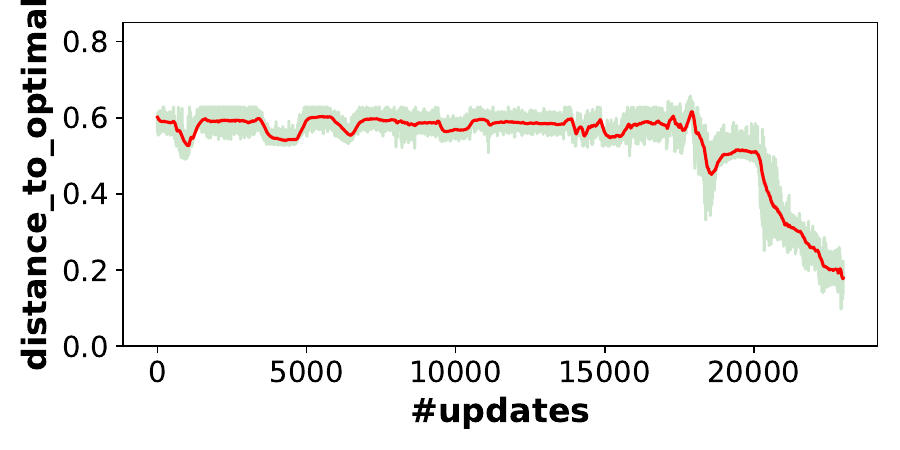} &  \includegraphics[width=0.45\linewidth]{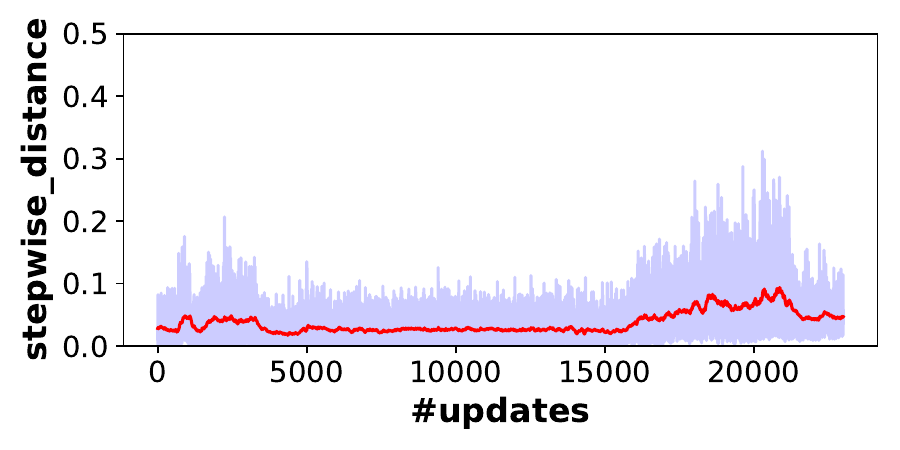}       
\end{tabular}
\caption{\textit{distance-to-optimal} and \textit{stepwise-distance} vs updates plots for DDPG algorithm in the Mountain Car continuous environment.}\label{ddpg_policy_evolution}\vspace*{0.5em}
\end{figure}

\begin{figure}[h!]
\centering\vspace*{0em}
\setlength{\arrayrulewidth}{0.1mm}
\setlength{\tabcolsep}{0.0mm}
\begin{tabular}{ll}
\includegraphics[width=0.45\linewidth]{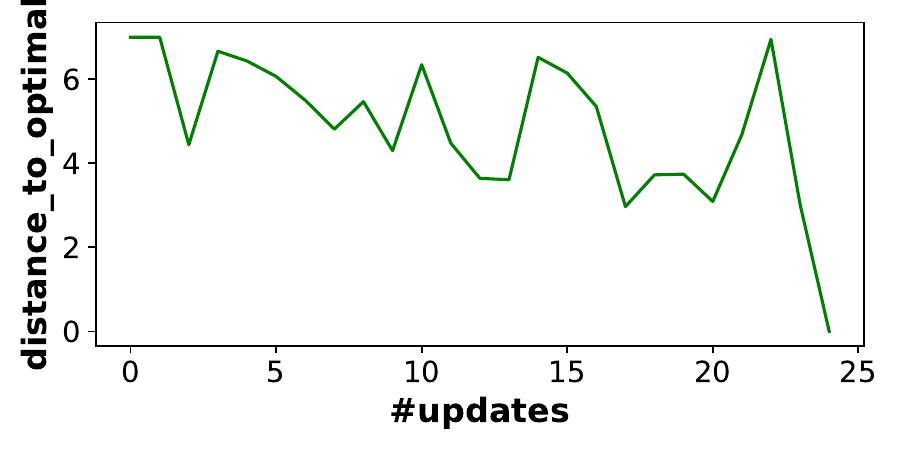} &  \includegraphics[width=0.45\linewidth]{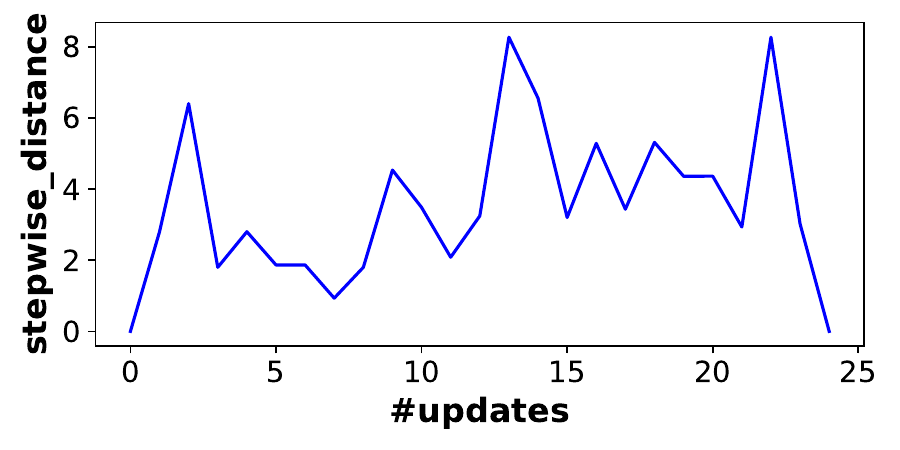} \\     
\includegraphics[width=0.45\linewidth]{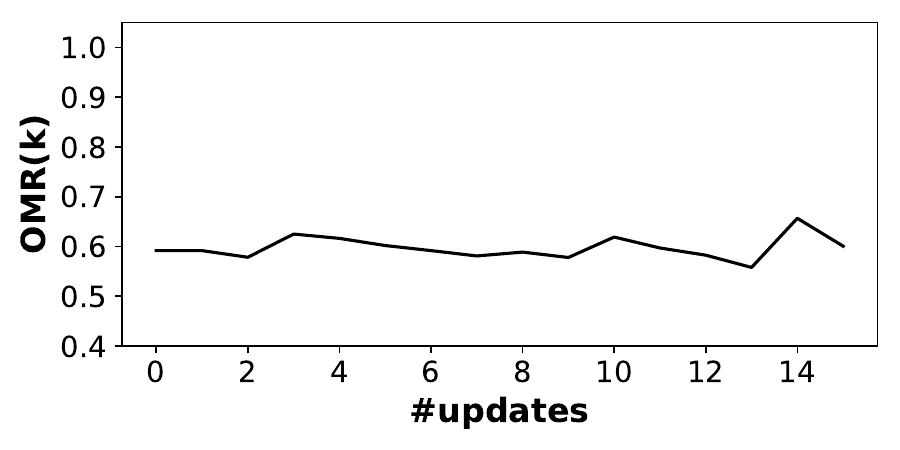} 
\end{tabular}
\caption{\textit{distance-to-optimal} and \textit{stepwise-distance} vs updates plots for PSRL algorithm in the 2D-Gridworld environment.}\label{psrl_policy_evolution}\vspace*{0.5em}
\end{figure}

\begin{figure}[h!]
\centering\vspace*{0em}
\setlength{\arrayrulewidth}{0.1mm}
\setlength{\tabcolsep}{0.0mm}
\begin{tabular}{ll}
\includegraphics[width=0.45\linewidth]{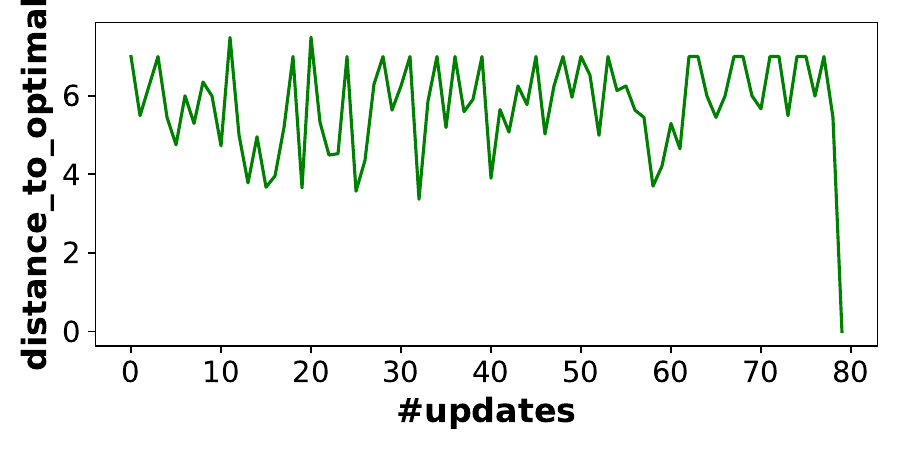} &  \includegraphics[width=0.45\linewidth]{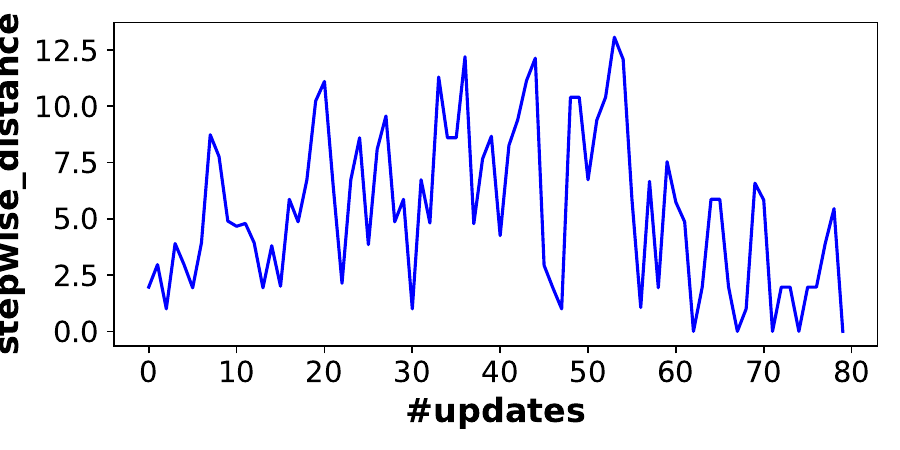} \\       
\includegraphics[width=0.45\linewidth]{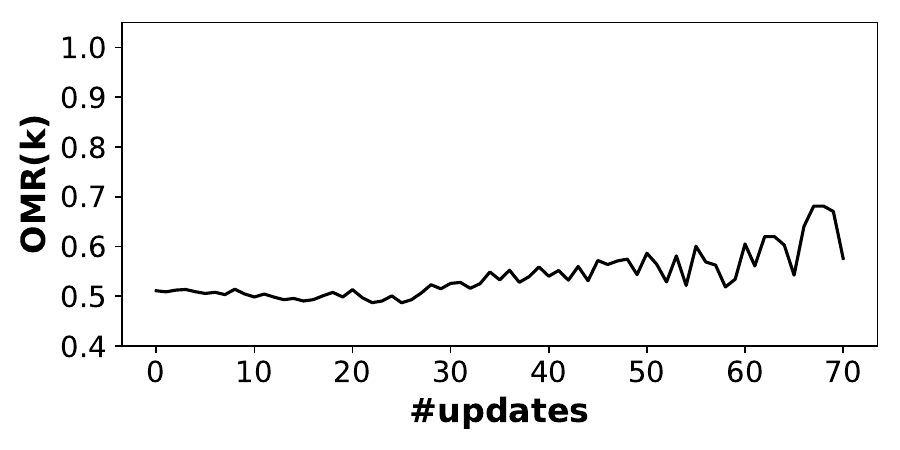} 
\end{tabular}
\caption{\textit{distance-to-optimal} and \textit{stepwise-distance} vs updates plots for UCRL algorithm in the 2D-Gridworld environment.}\label{ucrl_policy_evolution}\vspace*{0.5em}
\end{figure}

\begin{figure}[h!]
\centering\vspace*{0em}
\setlength{\arrayrulewidth}{0.1mm}
\setlength{\tabcolsep}{0.0mm}
\begin{tabular}{ll}
\includegraphics[width=0.45\linewidth]{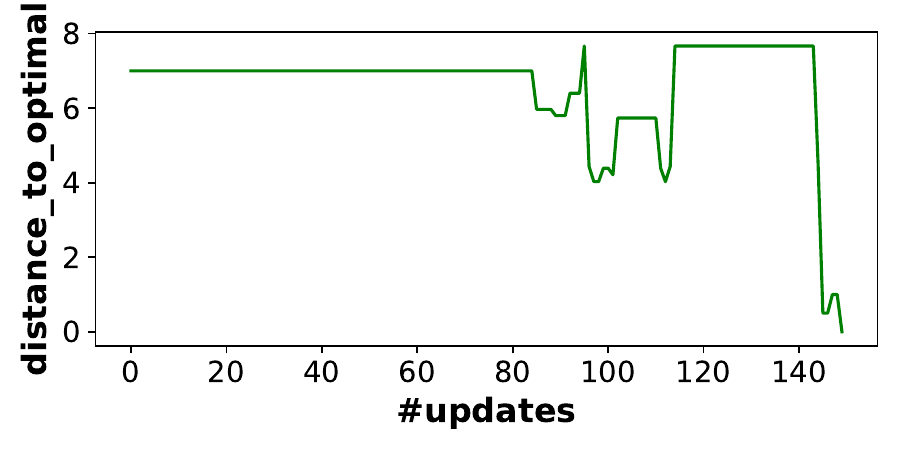} &  \includegraphics[width=0.45\linewidth]{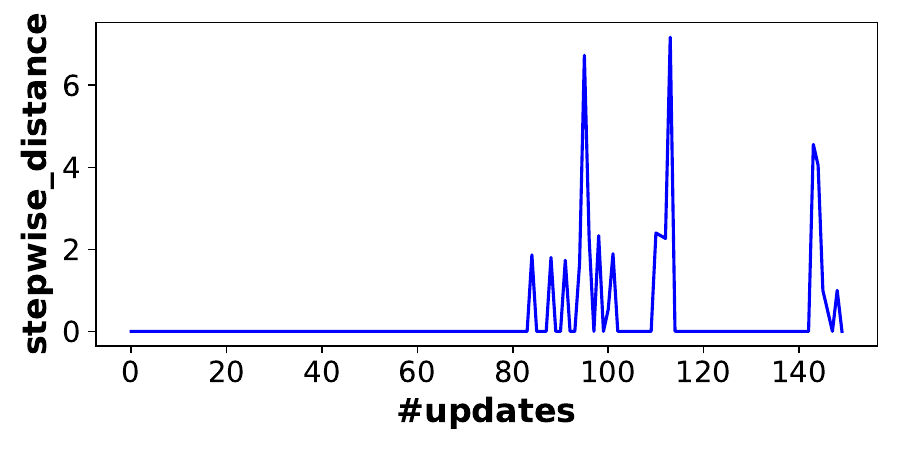} \\      
\includegraphics[width=0.45\linewidth]{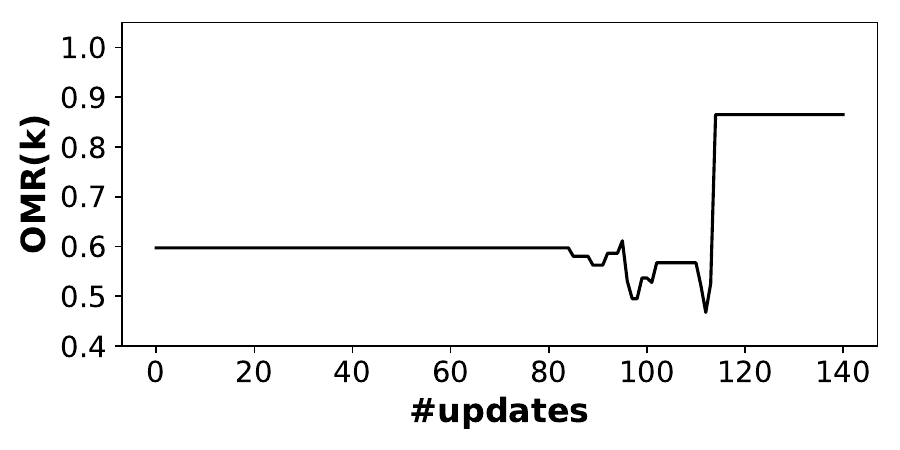} 
\end{tabular}
\caption{\textit{distance-to-optimal} and \textit{stepwise-distance} vs updates plots for DQN algorithm in the 2D-Gridworld environment.}\label{dqn_policy_evolution}\vspace*{-0.5em}
\end{figure}

In this section we present 2 dimensional versions of the policy trajectories in Figures~\ref{policy_evolution} and \ref{contin_policy_evolution}. These are \textit{distance-to-optimal} vs updates and \textit{stepwise-distance} vs. updates plots for selected algorithms (Figures~\ref{sac_policy_evolution} - \ref{dqn_policy_evolution}). Furthermore,  OMR($k)$ plots for algorithms used in the discrete MDP are presented. 

\clearpage


\end{document}